\documentclass[11pt]{article}
\usepackage{acl}
\usepackage{times}
\usepackage{latexsym}
\usepackage{graphicx}
\usepackage{amsmath,amssymb}
\usepackage{booktabs}
\usepackage{multirow}
\usepackage{microtype}
\usepackage{xspace}
\usepackage{enumitem}
\usepackage{subcaption}
\usepackage{inconsolata}
\usepackage{cleveref}
\usepackage[most]{tcolorbox}

\newtcolorbox{insightbox}{
  enhanced,
  colback=brown!20,          
  colframe=brown!90!black,  
  boxrule=1pt,              
  leftrule=1pt,             
  rightrule=1pt,            
  toprule=1pt,             
  bottomrule=1pt,          
  before skip=5pt,
  after skip=5pt,
  top=3pt,
  bottom=3pt,
}

\title{Context-Aware Decoding for Faithful Vision–Language Generation}

\author{Mehrdad Fazli, Bowen Wei, Ziwei Zhu \\
  Department of Computer Science, George Mason University \\
  Fairfax, VA 22030, USA \\
  \texttt{\{mfazli, bwei2, zzhu20\}@gmu.edu}}
  
\begin{document}
\maketitle
\begin{abstract}
Hallucinations—generating responses inconsistent with the visual input—remain a critical limitation of large vision-language models (LVLMs), especially in open-ended tasks such as image captioning and visual reasoning. In this work, we probe the layer-wise generation dynamics that drive hallucinations and propose a training-free mitigation strategy. Employing the Logit Lens, we examine how LVLMs construct next-token distributions across decoder layers, uncovering a pronounced \emph{commitment-depth gap}: truthful tokens accumulate probability mass on their final candidates earlier than hallucinatory ones. Drawing on this discovery, we introduce Context Embedding Injection (CEI), a lightweight method that harnesses the hidden state of the last input token—the \emph{context embedding}—as a grounding signal to maintain visual fidelity throughout decoding and curb hallucinations. Evaluated on the CHAIR, AMBER, and MMHal-Bench benchmarks (with a maximum token length of 512), CEI outperforms state-of-the-art baselines across three LVLMs, with its dynamic variant yielding the lowest overall hallucination rates. By integrating novel mechanistic insights with a scalable intervention, this work advances the mitigation of hallucinations in LVLMs.\footnote{This is a preprint under review.}
\end{abstract}

\section{Introduction}
Large vision–language models (LVLMs)~\citep{baivl,chen_shikra_2023,liu2023visual,internvl,dai_instructblip_2023,ye_mplug-owl_2024} have rapidly advanced general-purpose multimodal understanding by pairing strong vision encoders (e.g., CLIP~\citep{radford_learning_2021}) with large language models (e.g., LLaMA~\citep{touvron_llama_2023}). They excel at image captioning, visual question answering, and medical report generation~\citep{chen_vlp_2023,wu_multimodal_2023,hartsock_vision-language_2024}. Recent systems further scale in capability and robustness across diverse image resolutions and inputs, expanding their application~\citep{wang_qwen2-vl_2024,liu2024llavanext}. In spite of these strengths, LVLMs remain susceptible to \emph{hallucination}—producing text that deviates from the image—undermining reliability in applications that demand factuality and faithfulness, such as autonomous driving and medical report generation~\citep{keskar_evaluating_nodate,hartsock_vision-language_2024}.

Extensive research has sought to analyze hallucinations in LVLMs. Prior work attributes them to interacting factors, with analyses predominantly focusing on attention mechanisms. Studies highlight biased cross-attention to a few image patches~\cite{avisc, zhu_mitigating_2025}, attention drift during long generations~\cite{fazli_mitigating_2025, zhou_analyzing_2024}, and attention aggregation patterns such as “anchor/summary” tokens~\cite{huang_opera_nodate, zhang_seeing_2024} as key contributors. Despite these findings, \textit{the layer-wise progression of token prediction—how models accrue probability mass on their final decision set—remains unexplored.}


Meanwhile, various mitigation strategies have been proposed, including data optimization (e.g., negative samples), architectural enhancements (e.g., improved alignment modules), and decoding optimizations (e.g., contrastive schemes)~\citep{huang_survey_2025}. Inference-time interventions, such as contrastive decoding~\citep{vcd, m3id, zhu_ibd_2024} and attention calibration~\citep{fazli_mitigating_2025, liu_paying_2024}, have gained traction for their low cost and generalizability. However, \textit{these methods often achieve inconsistent performance in open-ended generative scenarios, such as image captioning with performance degrading as sequence length increases due to accumulated errors and drift from visual grounding~\citep{fazli_mitigating_2025}}.


To address these gaps, instead of characterizing hallucinations via post-hoc cues such as raw attention maps or dataset-level correlations, we apply Logit Lens~\cite{LogitLens2020}—a mechanistic interpretability technique—to directly track how token preferences form across decoder layers during decoding.
This analysis unveils a profound commitment-depth gap: truthful tokens stabilize probability mass on their final candidates earlier across decoder layers than hallucinatory ones, offering mechanistic insights into the roots of hallucination in LVLMs. Building on this discovery, we introduce Context Embedding Injection (CEI), a novel model-agnostic, training-free method that continually aligns the hidden states of generated tokens with a visual grounding signal—the context embedding—extracted from the initial input processing, thereby sustaining fidelity in open-ended generation. Our contributions are threefold:
\begin{itemize}
\item An empirical analysis, via the Logit Lens, demonstrating that truthful tokens consolidate predictions earlier across decoder layers than hallucinatory ones.
\item A novel mitigation method, Context Embedding Injection (CEI), for sustained visual alignment in open-ended generation.
\item Consistent improvements over baselines on generative hallucination benchmarks such as CHAIR, AMBER, and MMHal-Bench (GPT4-evaluated).
\end{itemize}

\section{Related Work}
\label{related_work}

\subsection{Large Vision-Language Models}
Large vision-language models (LVLMs) extend LLMs with visual inputs via a vision encoder (e.g., CLIP~\citep{radford_learning_2021}, ViT~\citep{dosovitskiy_image_2021}), an alignment module (e.g., linear projection~\cite{liu2023visual, liu2024llavanext} or Q-former~\citep{dai_instructblip_2023, minigpt4}), and an autoregressive LLM backbone (e.g., LLaMA~\citep{touvron_llama_2023}, Vicuna~\cite{zheng_judging_2023}). Recent families~\citep{liu2024llavanext, wang_qwen2-vl_2024, ye_mplug-owl_2024} scale data/model size and improve visual tokenization and positional fusion (e.g., multi-scale inputs and M-RoPE~\citep{liu2024llavanext, wang_qwen2-vl_2024}), yielding broad gains across captioning and VQA. Despite these advances, LVLMs remain prone to hallucination, limiting reliability in safety-critical settings~\citep{bai_hallucination_2025}.

\subsection{Hallucination Mitigation in LVLMs}
LVLM hallucinations are outputs that deviate from the image (e.g., fabricated objects/attributes/relations), commonly linked to linguistic priors, dataset bias, and modality misalignment~\citep{bai_hallucination_2025, pope, liu_survey_2024}. Existing mitigation methods span fine-tuning~\citep{gunjal_detecting_2024, jiang_hallucination_2024, kim_exposing_2023}, post-hoc correction~\citep{yin_woodpecker_2023, zhou_analyzing_2024}, and decoding-time approaches~\citep{vcd, opera, suo_octopus_2025, fazli_mitigating_2025, an_mitigating_2025, yang_nullu_2025}. Decoding-time methods are especially attractive since they require no retraining; prominent examples include contrastive decoding variants that compare original vs. perturbed inputs to promote visual grounding (e.g., VCD~\citep{vcd}, M3ID~\citep{m3id}, IBD~\citep{zhu_ibd_2024}) as well as attention-centric calibration of cross-modal interactions (e.g., AGLA~\citep{an_mitigating_2025}, CAAC~\citep{fazli_mitigating_2025}). In contrast, our method intervenes in the embedding space, continually aligning token representations with a visual grounding signal. For a more comprehensive discussion of related work and additional references, see Appendix~\ref{app:related_work}.

\section{Preliminary Insights} \label{sec:prelims}


To effectively mitigate hallucinations in large vision-language models (LVLMs), a nuanced understanding of the underlying generation dynamics is essential. A key question is: \emph{what internal signals and structural patterns distinguish truthful generations from hallucinatory ones?} Motivated by recent advances in mechanistic interpretability, we apply targeted mechanistic probes to decoder layers to address this question, revealing the internal signals responsible for divergence from visual information. In Section~\ref{sec:first_to_know}, we identify a reusable grounding signal embedded in the initial decoding step, and in Section~\ref{sec:analysis}, we uncover a systematic commitment-depth gap between truthful and hallucinatory tokens—insights that directly inform our subsequent intervention strategy.

\subsection{Initial Decoding Step as a Grounding Signal} \label{sec:first_to_know}

\begin{figure}
    \centering
    \includegraphics[width=0.8\linewidth]{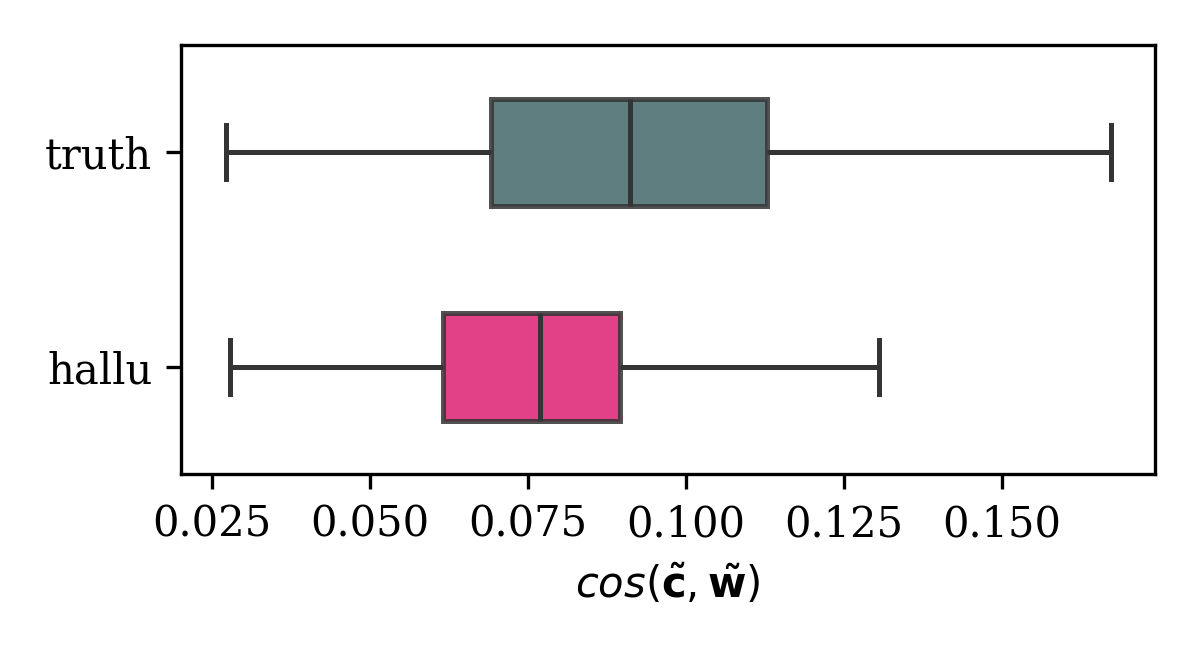}
    \caption{\textbf{Context embedding semantic alignment}. Box plot of centered cosine similarities $\cos(\tilde{\mathbf{c}}, \tilde{\mathbf{w}})$ between the centered context embedding $\tilde{\mathbf{c}}$ and centered target token embedding $\tilde{\mathbf{w}}$. Blue and orange boxes denote truthful and hallucinatory tokens respectively.}
    \label{fig:centered-cosine}
    \vspace{-10pt}
\end{figure}

We hypothesize that the hidden state of the last input token---the \emph{context embedding}---encodes query-aligned visual information, steering the logit distribution of the first generated token toward image-grounded content while serving as a reusable anchor to sustain visual fidelity throughout decoding.

To test this, we evaluate the semantic alignment between the context embedding and target tokens using the generative captioning split of the AMBER dataset~\citep{amber}, which provides annotations for truthful and hallucinatory objects. Using InstructBLIP~\citep{dai_instructblip_2023}, we generate captions, extract target tokens, and compute centered cosine similarities $\cos(\tilde{\mathbf{c}}, \tilde{\mathbf{w}})$ between the centered context embedding $\tilde{\mathbf{c}}$ and centered token embeddings $\tilde{\mathbf{w}}$ (following \citealp{muennighoff_mteb_2023} to account for anisotropy).


As shown in \Cref{fig:centered-cosine}, truthful tokens exhibit significantly higher mean cosine similarities than hallucinatory ones (95\% CI excluding zero), confirming the context embedding's inclination toward grounded content. This aligns with \citet{zhao2024first}, who leverage first-token hidden states for detecting hallucinations via lightweight probes. Together, these findings indicate that \emph{the context embedding is intrinsically more semantically aligned with visually supported tokens than with hallucinated ones.}

\subsection{Layer-wise Commitment in Truthful vs. Hallucinatory Tokens} \label{sec:analysis}

Prior research has shown that language models determine function words within mid-layers, maintaining stable predictions thereafter, while content words involve ongoing adjustments in later layers~\cite{zhu_ibd_2024}. Motivated by this, we ask if a similar disparity in commitment depth underlies the generation of truthful and hallucinatory tokens. Specifically, we investigate \emph{if truthful tokens accrue probability mass on their final decision set earlier than hallucinatory ones.}

Similar to the experimental pipeline from ~\Cref{sec:first_to_know}---which leverages the AMBER generative split~\citep{amber} for caption generation and target token identification---we perform, for each truthful or hallucinatory token, a forward pass on the prefix up to each target token to extract intermediate hidden states $h_\ell$ at layer $\ell$. We then apply the Logit Lens~\citep{LogitLens2020} to map these states to vocabulary distributions $p_\ell = \sigma(h_\ell E^\top)$, where $E$ is the embedding matrix.




\begin{figure}[t]
  \centering
  \begin{subfigure}[b]{0.24\textwidth}
    \centering
    \includegraphics[width=\linewidth]{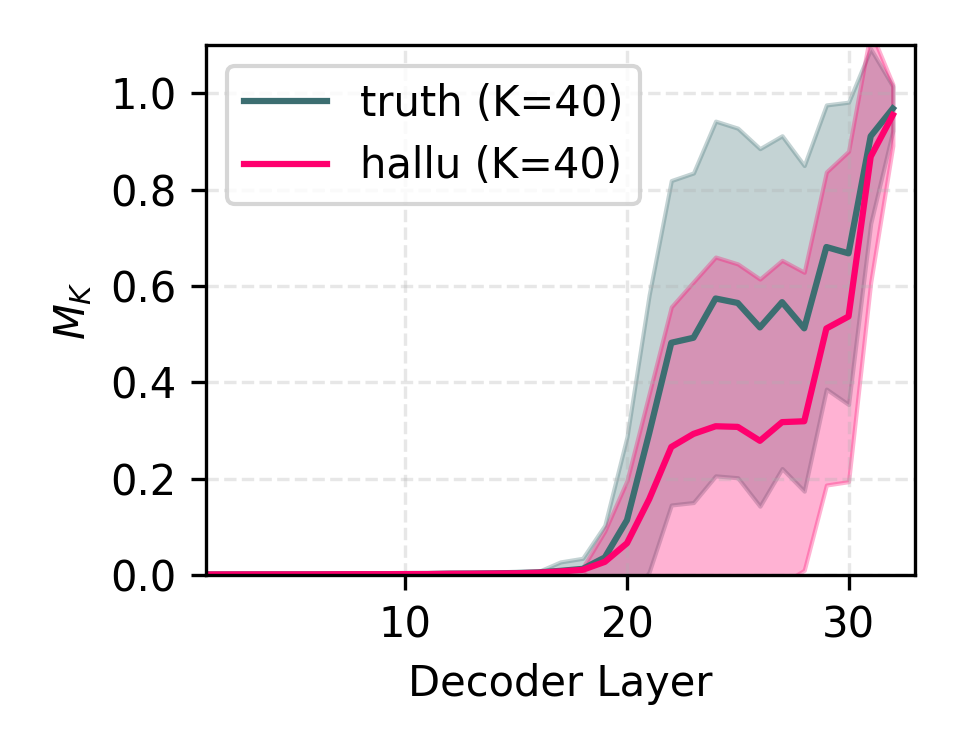}
    \vspace{-20pt}
    \caption{}
    \label{}
  \end{subfigure}\hfill%
  \begin{subfigure}[b]{0.24\textwidth}
    \centering
    \includegraphics[width=\linewidth]{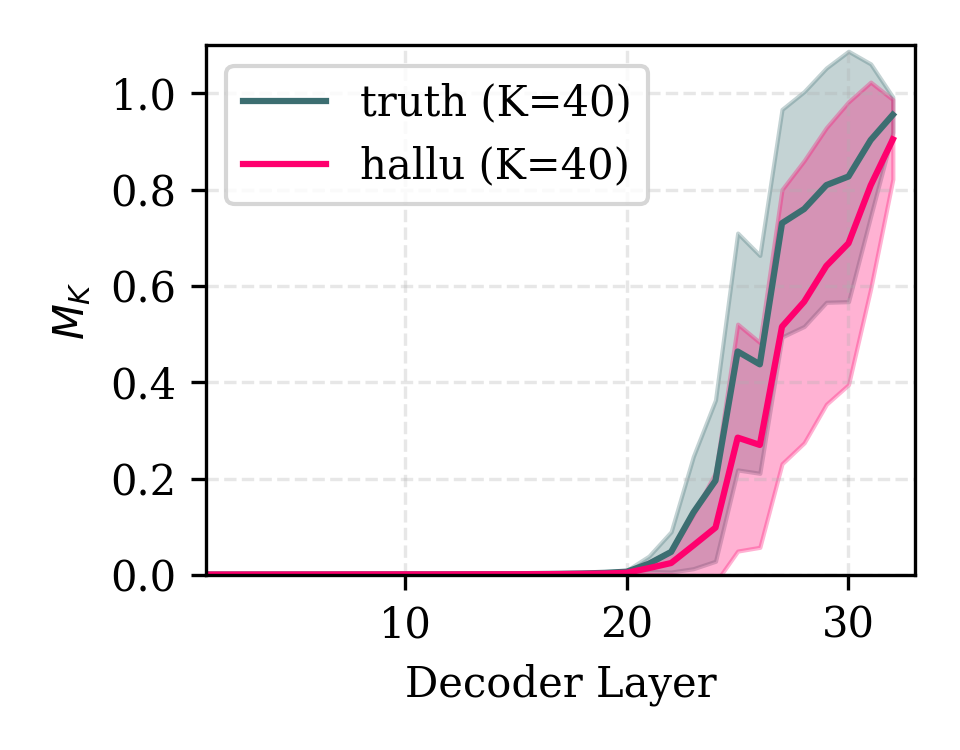}
    \vspace{-20pt}
    \caption{}
    \label{}
  \end{subfigure}\hfill%
  \caption{\textbf{Layer-wise commitment curves.} Top-$K$ probability mass, $M_K(\ell)$, of truthful (blue) and hallucinatory (red) tokens across decoder layers, with shaded bands indicating one standard deviation: (a) InstructBLIP and (b) LLaVA-1.5.
  }
  \vspace{-10pt}
  \label{M_K_plots}
\end{figure}

\begin{figure}[t]
  \centering
  \begin{subfigure}[b]{0.24\textwidth}
    \centering
    \includegraphics[width=\linewidth]{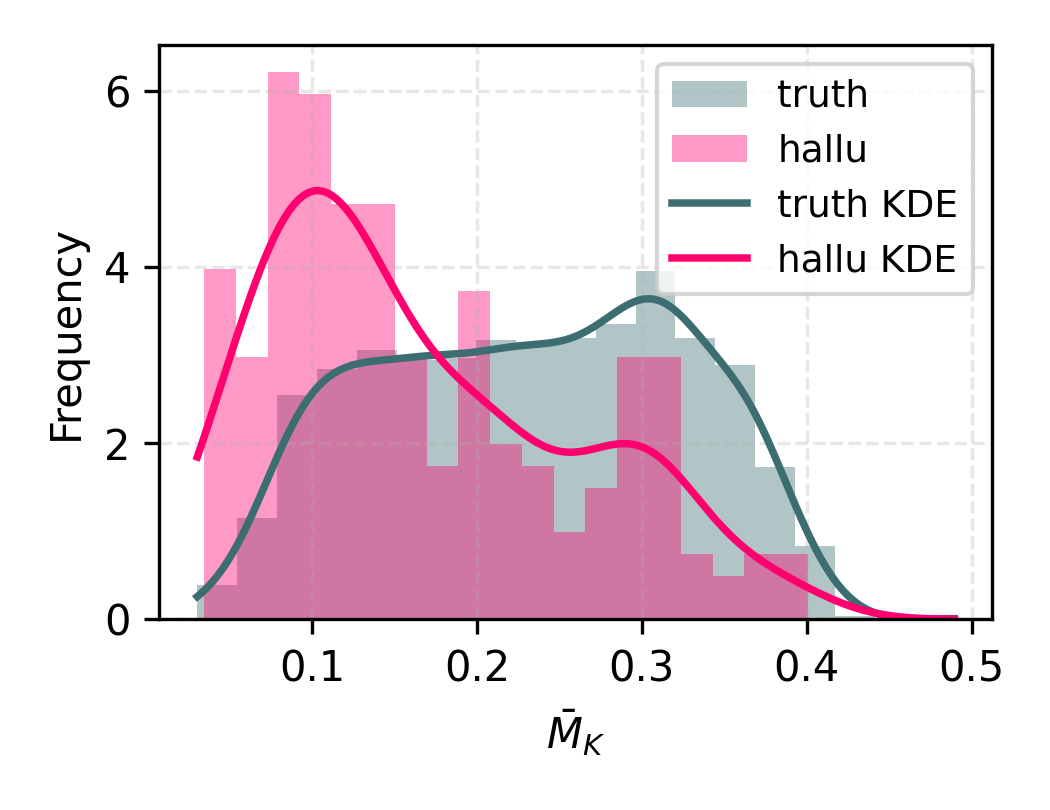}
    \vspace{-20pt}
    \caption{}
    \label{}
  \end{subfigure}\hfill%
  \begin{subfigure}[b]{0.24\textwidth}
    \centering
    \includegraphics[width=\linewidth]{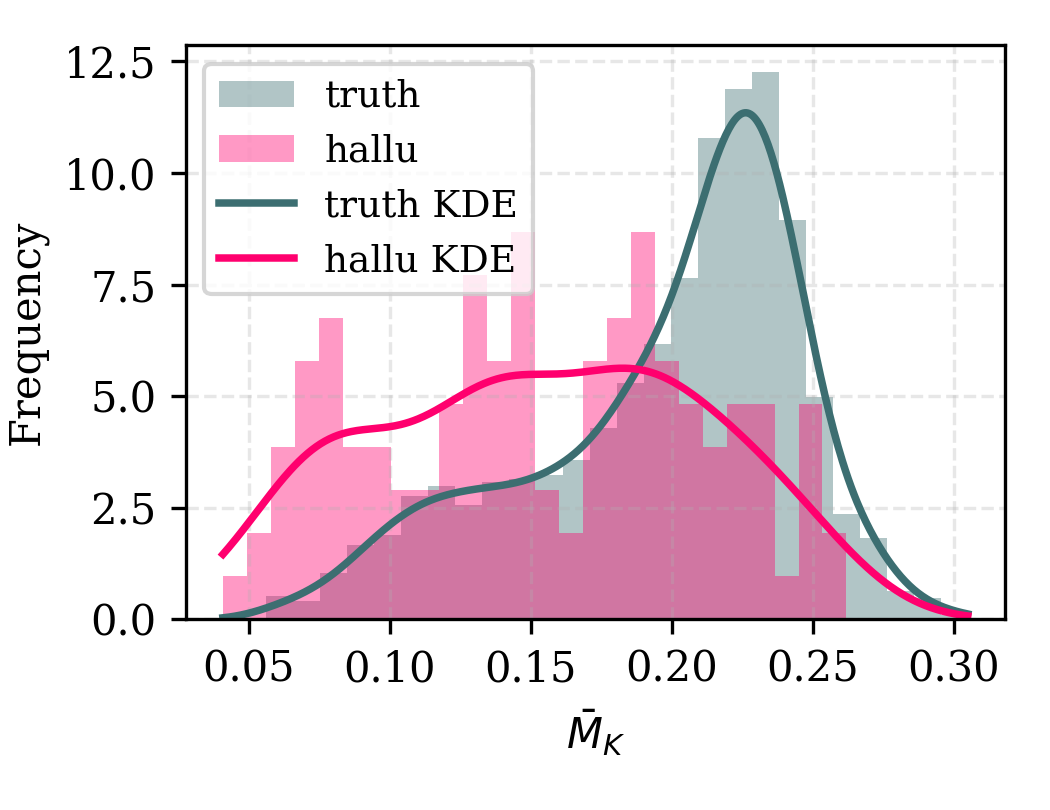}
    \vspace{-20pt}
    \caption{}
    \label{}
  \end{subfigure}\hfill%
  \caption{\textbf{Average confidence in decision set.} Histogram of mean top-$K$ probability mass, $\bar{M}_K$, of truthful (blue) and hallucinatory (red) tokens for (a) InstructBLIP and (b) LLaVA-1.5 ($K=40$).
  }
  \vspace{-10pt}
  \label{bar_M_K_plots}
\end{figure}

Let $S_K$ be the \textit{final decision set} defined as top-$K$ tokens under the distribution $p_L$. The layer-wise top-$K$ probability mass is defined as:
\begin{equation}
    M_K(\ell) = \sum_{w \in S_K} p_{\ell}(w),
\end{equation}
which measures how much probability layer $\ell$ assigns to the final decision set. To quantify the overall confidence of the model on the final decision set, we compute the layer-wise average top-$K$ mass:
\begin{equation}
    \bar{M}_K = \frac{1}{L} \sum_{\ell=1}^L M_K(\ell).
\end{equation}

Across InstructBLIP~\citep{dai_instructblip_2023}, LLaVA~\citep{liu2023visual}, and LLaVA-NeXT~\citep{liu2024llavanext}, truthful tokens show earlier commitment (higher $M_K(\ell)$ in mid-to-late layers, $\ell \approx 20$--$30$; see \Cref{M_K_plots}), with robust separation for $K \in [20, 80]$ (\Cref{app:context-embedding-analysis}). 


Furthermore, the distributions of mean top-$K$ mass $\bar{M}_K$ differ markedly between truthful and hallucinatory tokens (\Cref{bar_M_K_plots}): hallucinatory ones exhibit lower values, signaling delayed probability consolidation on final candidates, while truthful tokens accrue confidence earlier across layers, yielding stronger, more stable commitments.


\begin{insightbox}
Truthful tokens commit earlier and accrue higher cumulative confidence on their final decision set, revealing a clear commitment-depth gap between truthful and hallucinatory generations.
\end{insightbox}


\section{Proposed Method}
\begin{figure*}[t]
    \centering
    \includegraphics[width=0.7\linewidth]{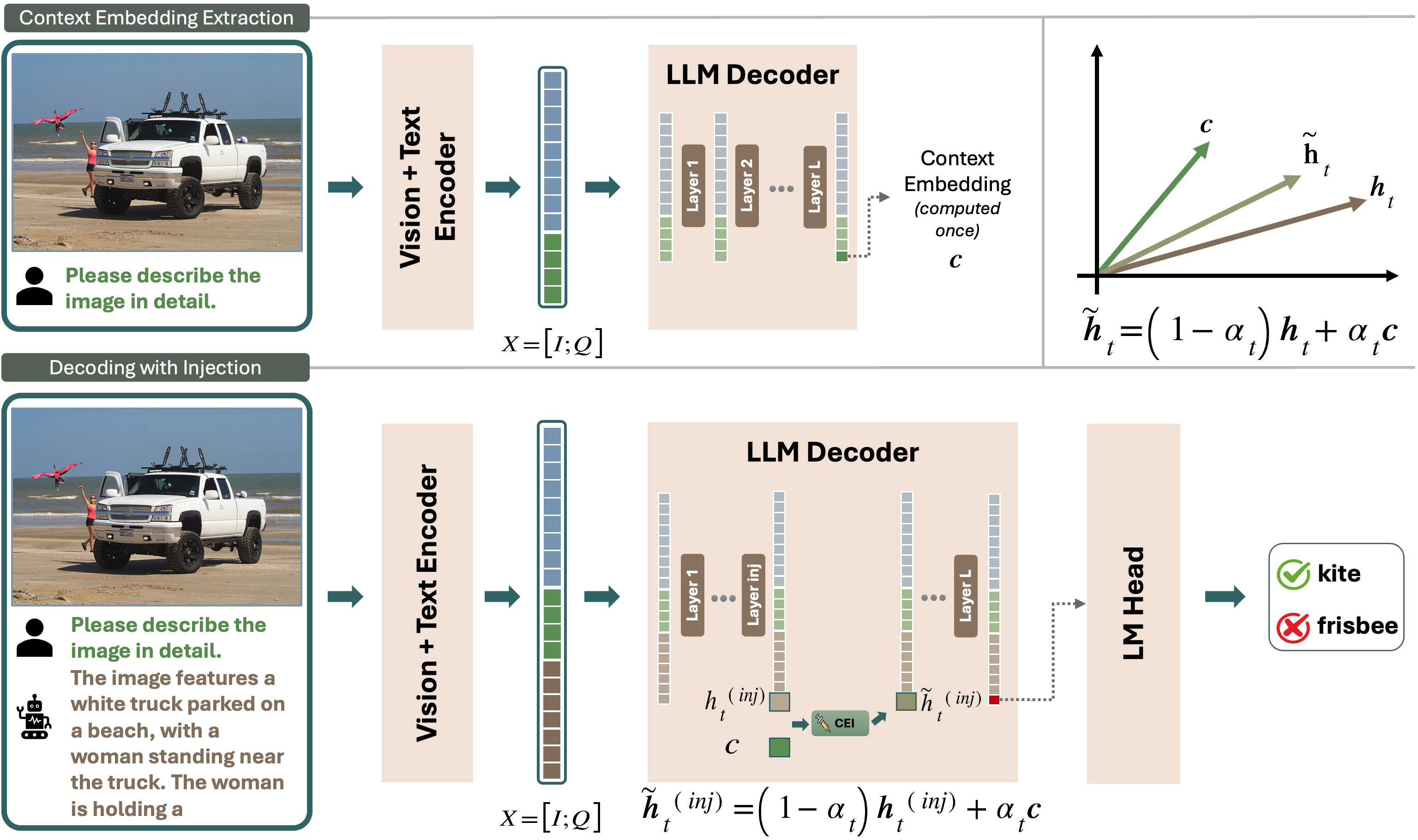}
    \caption{\textbf{CEI overview.} An initial forward pass over the image--prompt input extracts a fixed \emph{context embedding}~$\mathbf{c}$ from the final decoder layer at the last prompt position (top). During autoregressive decoding, this pre-computed signal is injected at a chosen decoder layer via a weighted average mechanism, guiding generation toward the original image-conditioned context and improving visual grounding (bottom). Blue, green, and brown squares denote the embeddings of image tokens, prompt tokens, and previously generated tokens, respectively.}
    \label{fig:CEI_overview}
    \vspace{0pt}
\end{figure*}

We propose CEI, a training-free decoding intervention that reuses the context embedding to maintain outputs visually grounded (\Cref{fig:CEI_overview}). CEI extracts an image-conditioned context embedding before generation and re-injects it into the decoder during autoregressive decoding to anchor token predictions to the original visual input. \Cref{sec:inference_lvlm} fixes the inference notation; \Cref{sec:static-cei} introduces \emph{Static CEI}, which injects this fixed context into the last-input hidden state at a chosen layer with a constant weight \(\alpha\). \Cref{sec:dynamic-cei} presents \emph{Dynamic CEI}, where \(\alpha\) is modified per token from \(\bar M_K\) when commitment is weak.

\subsection{Inference in Large Vision-Language Models} \label{sec:inference_lvlm}

The inference process in LVLMs comprises two main stages: \emph{input preparation} and \emph{text generation}. These stages facilitate autoregressive text generation conditioned on multimodal inputs.

\paragraph{Input preparation.}
The processor module transforms the raw inputs into a unified embedding space suitable for the autoregressive language decoder.

A vision encoder $E_v$ extracts features from the input image $I$, which are then projected into the textual embedding space via an alignment module (e.g., linear projection or Q-former) to yield visual tokens $V = [v_1, \dots, v_{N_v}] \in \mathbb{R}^{N_v \times d}$, where $d$ is the embedding dimension.
The text tokenizer $E_t$ embeds the input sequence $q = [P; y_{<t}] = [q_1, \dots, q_{N_q}]$, where $P$ is the prompt and $y_{<t}$ are previously generated tokens, producing $Q = E_t(q) \in \mathbb{R}^{N_q \times d}$.

The multimodal input is then formed by concatenation:
\begin{equation}
    X = [V; Q] \in \mathbb{R}^{(N_v + N_q) \times d}.
\end{equation}

\paragraph{Text generation.}
The decoder, with $L$ layers, processes $X$ autoregressively. At step $t$, hidden states evolve as $H^{(\ell)} = f^{(\ell)}(H^{(\ell-1)})$ where $\ell = 1, \dots, L$, $f^{(\ell)}$ being the $\ell$-th transformer layer, and $H^{(0)} = X$.
The final hidden state at the last position, $h^{(L)}_t \in \mathbb{R}^d$, aggregates contextual information from the entire sequence for predicting the next token.
This state is projected back to the vocabulary space via the unembedding matrix $E^\top \in \mathbb{R}^{d \times |\mathcal{V}|}$: to yield logits:
\begin{equation}
    z_t = h^{(L)}_t E^\top,
\end{equation}
followed by a softmax results in a probability distribution over the vocabulary $\mathcal{V}$:
\begin{equation}
    p_\theta(y_t \mid x, p, y_{<t}) = \mathrm{softmax}(z_t).
\end{equation}
A decoding strategy (e.g., greedy selection, nucleus sampling, or beam search) selects the next token $\hat{y}_t$, which is appended to the sequence for the subsequent iteration. This autoregressive process continues until a termination criterion is met.

\subsection{Static Context Embedding Injection}
\label{sec:static-cei}

Building on the insights from Section~\ref{sec:first_to_know}, we propose Context Embedding Injection (CEI), a training-free method to enhance visual grounding in LVLMs during autoregressive generation. We define \emph{context embedding} as the hidden state at the last prompt token in the final decoder layer, which directly yields the logits for the first generated token. This representation preserves the full information of the initial logit distribution while being amenable to blending with subsequent hidden states. CEI leverages context embedding in subsequent decoding steps to reinforce visual alignment throughout the generation process.

Let the prompt be tokenized into $N_p$ tokens. The multimodal input for the initial forward pass (before generation) is then $X_0 = [V; P] \in \mathbb{R}^{(N_v + N_p) \times d}$. Prior to generation, we perform an initial forward pass on the multimodal input $X$ to extract the context embedding $\mathbf{c}$ from the last decoder layer at the last input position $Q$:
\begin{equation}
    \mathbf{c} = h^{(L)}_{N_v+N_p}.
\end{equation}
During decoding, we inject this fixed signal at a chosen layer $\ell_{\mathrm{inj}} \in {1, \dots, L}$ using a constant mixing weight \(\alpha\in[0,1]\). For each step \(t\),
\begin{equation}
    \tilde{h}^{(\ell_{\mathrm{inj}})}_{t}
= (1-\alpha)\,h^{(\ell_{\mathrm{inj}})}_{t}
\;+\; \alpha\,\mathbf{c}.
\end{equation}
The blended hidden state $\tilde{h}_t^{(\ell{\mathrm{inj}})}$ is thereby aligned more closely with the context embedding. This alignment propagates through subsequent layers, influencing the next-token distribution and promoting fidelity to the input image. Static CEI computes \(\mathbf{c}\) once (before decoding) and applies uniform injection across all tokens, incurring only a lightweight vector blending operation without requiring additional training. The hyperparameters $\alpha$ and $\ell_{\mathrm{inj}}$ are tuned for each LVLM. 

While Static CEI effectively reinforces visual grounding, it applies a uniform injection across all decoding steps. In practice, however, not all tokens are equally image-dependent: function tokens primarily support linguistic fluency, whereas content tokens (e.g., objects, attributes, actions) are more tightly coupled to the visual input. This mismatch motivates a dynamic formulation that adapts the injection strength at the token level.

\subsection{Dynamic Context Embedding Injection}
\label{sec:dynamic-cei}

The analysis in Section~\ref{sec:analysis} reveals that hallucinatory tokens exhibit lower mean top-$K$ probability mass $\bar{M}_K$ compared to truthful ones, indicating delayed commitment and a higher risk of deviation from visual grounding. We leverage this signal to extend static CEI by dynamically modulating the injection weight $\alpha_t$ per generation step $t$.

At each step $t$, we first compute $\bar{M}_K$ via a probe forward pass as explained in Section~\ref{sec:analysis}. We then map it to $\alpha_t$ using a half-cosine schedule:
\begin{equation}
    \alpha_t = \min( \alpha_{max} \cos(\frac{\pi}{2}.\frac{\bar{M}_K}{\beta}), 0).
\end{equation}
This schedule ensures low $\bar{M}_K$ values that signal high hallucination risk, enforce stronger injection (higher $\alpha$) for better alignment. While risk decreases rapidly as $\bar{M}_K$ rises; $\beta$ also indicates the cutoff where injection tapers off. The blended hidden state is then formed as in static CEI, but with the adaptive $\alpha_t$:
\begin{equation}
    \tilde{h}^{(\ell_{\mathrm{inj}})}_{t}
= (1-\alpha_t)\,h^{(\ell_{\mathrm{inj}})}_{t}
\;+\; \alpha_t\,\mathbf{c}.
\end{equation}
A second forward pass is subsequently performed with the adjusted injection weight to predict the next token. Dynamic CEI thus adapts the injection online based on generation dynamics, enhancing robustness without requiring additional training.

\section{Experiments and Results}
\subsection{Experimental Setup} \label{exp-setup}

\noindent\textbf{Models.}
Following the experimental protocols used in recent works on LVLM hallucination, we evaluate the proposed CEI method on three widely used 7B-parameter LVLMs: InstructBLIP~\citep{dai_instructblip_2023}, LLaVA-1.5~\citep{liu2023visual}, and LLaVA-NeXT~\citep{liu2024llavanext}. The CEI framework is designed to be model-agnostic, enabling integration with diverse LVLMs. Detailed experimental configurations and implementation specifics are provided in \Cref{app:exp-details}.

\paragraph{Benchmarks.}
We focus on generative benchmarks that assess faithfulness in open-ended generation. Specifically, we adopt the CHAIR~\citep{rohrbach2019object}, AMBER~\citep{amber}, and MMHal-Bench~\cite{sun_aligning_2023} benchmarks. We set the maximum number of new tokens to 512 for all benchmarks to allow uninterrupted, free-form generations, as hallucinations often occur in the later stages of decoding.

\paragraph{Metrics.}
For CHAIR, we report \(\mathrm{CHAIR}_i\) and \(\mathrm{CHAIR}_s\), which quantify object hallucinations at the instance and sentence levels, respectively. For AMBER, we report \(\mathrm{CHAIR}\) and \(\mathrm{HAL}\), which measure hallucination frequency and response-level occurrence, along with \(\mathrm{COVER}\), which evaluates completeness and informativeness. Lower \(\mathrm{CHAIR}\)/\(\mathrm{HAL}\) and higher \(\mathrm{COVER}\) indicate better grounding. For MMHal-Bench, we report Score and HalRate which measure the quality of the responses and hallucination rates.

\paragraph{Baselines.}
We compare CEI against five training-free hallucination mitigation techniques. Contrastive decoding methods include VCD~\citep{vcd}, AvisC~\citep{avisc}, and M3ID~\citep{m3id}, which leverage contrastive strategies, alongside OPERA~\citep{huang_opera_nodate}, a beam-search variant penalizing overconfident tokens for visual grounding, and CAAC~\citep{fazli_mitigating_2025}, an attention-calibration approach.

\paragraph{Implementation Details.}
Baseline hyperparameters follow settings from their original publications for consistency. For CEI, the critical hyperparameters of dynamic CEI, $\alpha_{\max}$ and $\beta$, govern the mapping from $\bar{M}_K$ to $\alpha$, and are tuned per LVLM, yielding $\alpha_{\max} = 0.4, 0.25, 0.17$ and $\beta = 0.7, 0.55, 0.35$ for InstructBLIP, LLaVA-1.5, and LLaVA-NeXT, respectively. The injection layer is fixed at 10, as experimental analysis indicated that layers 10–15 yield the most substantial improvements in faithfulness. More implementation details in \Cref{app:exp-details}. 

\subsection{Evaluation Results}
\paragraph{CHAIR.}
The Caption Hallucination Assessment with Image Relevance (CHAIR) benchmark~\citep{rohrbach2019object} quantifies object hallucinations in image captioning by comparing objects mentioned in generated captions against ground-truth annotations from the MSCOCO 2014 dataset~\citep{lin_microsoft_2015}. We report two standard metrics: $\mathrm{CHAIR}_i$, the proportion of hallucinated objects among all mentioned objects, and $\mathrm{CHAIR}_s$, the fraction of captions containing at least one hallucinated object, with lower values indicating superior performance. Following established protocols~\citep{huang_opera_nodate}, we evaluate on 500 randomly selected images from the MSCOCO validation set, prompting models with ``Please describe this image in detail." To capture hallucinations that often emerge in extended sequences, we set the maximum new tokens to 512, allowing uninterrupted generation.

As summarized in \cref{tab:chair_results}, both Static CEI and Dynamic CEI reduce object hallucination on \(\mathrm{CHAIR}_i\) and \(\mathrm{CHAIR}_s\) for nearly all metrics comparing to the baseline methods while Dynamic CEI attains the lowest \emph{average} CHAIR across models. The overall improvements of Dynamic over Static CEI align with its risk-aware adaptation, which adjusts intervention the commitment depth. Overall, these results indicate that reusing the pre-generation context to steer decoding is effective for curbing object hallucinations in image captioning.

\paragraph{AMBER.}
AMBER~\citep{amber} provides an LLM-free evaluation of LVLM hallucination. We use its \emph{generative} split, where models produce free-form descriptions scored by three metrics: \(\mathrm{CHAIR}\) (frequency of objects mentioned but not present), \(\mathrm{HAL}\) (fraction of responses containing any hallucination across objects/attributes/relations), and \(\mathrm{COVER}\) (fraction of image-grounded objects that are mentioned). Lower \(\mathrm{CHAIR}/\mathrm{HAL}\) and higher \(\mathrm{COVER}\) indicate better faithfulness and completeness. We adopt the official prompts and scoring and set \emph{max new tokens} to 512 to allow uninterrupted generations.

\Cref{tab:amber_results} reports results on AMBER's generative task. As shown, \emph{Dynamic CEI} consistently achieves the lowest hallucination rates across models, outperforming all baseline methods on both \(\mathrm{CHAIR}\) and \(\mathrm{HAL}\) while maintaining a comparable \(\mathrm{COVER}\) score, indicating improved faithfulness without sacrificing completeness. Despite its simplicity, \emph{Static CEI} also surpasses most baselines and performs comparably with \emph{CAAC}, the strongest attention-based mitigation method. In contrast, contrastive decoding approaches such as \emph{VCD} and \emph{M3ID} attain higher \(\mathrm{COVER}\) values—reflecting broader descriptive coverage—but exhibit substantially elevated hallucination rates, undermining their overall reliability. These results demonstrate that CEI offers a balanced and robust mitigation strategy for open-ended generation tasks.

\begin{table}[t]\small
  \centering
  \caption{Performance on CHAIR Benchmark}
  \vspace{-5pt}
  \setlength{\tabcolsep}{4pt}
  \begin{tabular}{lcc|cc|cc}
    \toprule
    \multirow{2}{*}{Method} & \multicolumn{2}{c|}{LLaVA-1.5} & \multicolumn{2}{c|}{InstructBLIP} & \multicolumn{2}{c}{LLaVA-NeXT} \\
     & \(\mathrm{C}_s\)$\downarrow$ & \(\mathrm{C}_i\)$\downarrow$ & \(\mathrm{C}_s\)$\downarrow$ & \(\mathrm{C}_i\)$\downarrow$ & \(\mathrm{C}_s\)$\downarrow$ & \(\mathrm{C}_i\)$\downarrow$ \\
    \midrule
    base & 55.2 & 17.6 & 55.6 & 16.6 & 33.0 & 9.4 \\
    + OPERA     & 44.6 & 12.8 & 46.4 & 14.2 & 39.4 & 11.8 \\
    + VCD       & 57.8 & 16.3 & 60.8 & 17.9 & 41.6 & 9.9 \\
    + AvisC     & 60.4 & 17.2 & 71.0 & 20.1 & 34.8 & 9.3 \\
    + M3ID      & 56.2 & 16.4 & 72.8 & 21.1 & 42.0 & 12.4 \\
    + CAAC      & 39.2 & \textbf{10.4} & 37.4 & 10.8 & 30.6 & 8.1 \\
    \midrule
    + Stat. CEI       & \underline{35.2}  & 13.8  & \underline{32.4}  & \textbf{8.5}   & \textbf{26.0}  & \underline{7.6}\\
    + Dyn. CEI       & \textbf{34.0}  & \underline{10.9}  & \textbf{32.2}  & \underline{8.7}   & \underline{26.4}  & \textbf{7.3}\\
    \bottomrule
  \end{tabular}
  \label{tab:chair_results}
  \vspace{-10pt}
\end{table}

\begin{table}[t]\small
  \centering
  \caption{Performance on AMBER Benchmark Across Different LVLMs}
  
  \vspace{-3pt}
  \begin{tabular}{l|ccc}
    \toprule
     & CHAIR$\downarrow$ & HAL$\downarrow$ & COVER$\uparrow$ \\
    
    \midrule
    InstructBLIP    & 12.8      & 53.5              & 52.7 \\
    + OPERA         & 9.7       & 40.5              & 51.2 \\
    + VCD           & 10.8      & 46.6              & \textbf{53.4} \\
    + M3ID          &  10.4     & 47.3              & 51.7\\
    + AvisC         & 10.1      & 46.8              & 51.2\\
    + CAAC          & 7.0       & \underline{30.9}     & \underline{51.9}\\
    \midrule
    + Stat. CEI     & \underline{6.1}   & 31.7          & \textbf{53.4} \\
    + Dyn. CEI      & \textbf{5.6}      & \textbf{30}   & \underline{51.9} \\
    \midrule
    \midrule
    LLaVA-1.5           & 11.3              & 48.1          & 50.4 \\
    + OPERA             & 7.3               & 29.5          & 47.5 \\
    + VCD               & 8.2               & 37.3          & 51.9\\
    + M3ID              & 7.2               &  41.4         & \textbf{57.3}\\
    + AvisC             & 11.0              & 48.0          & \underline{52.5} \\
    + CAAC              & \underline{6.0}   & \textbf{25.0} & 48.7 \\
    \midrule
    + Stat. CEI         & 6.4               & \underline{27.3}      & 48.6\\
    + Dyn. CEI          & \textbf{5.9}      & \textbf{25.0}         & 48.1 \\
    \midrule
    \midrule
    LLaVA-NeXT          & 9.3               & 51.3              & 60.6 \\
    + OPERA             & -                 & -                 & -  \\
    + VCD               & 10.5              & 57.2              & \textbf{63.5} \\
    + M3ID              & 12.4              & 59.8              & \underline{61.4} \\
    + AvisC             & 9.2               & 50.4              & 61.1 \\
    + CAAC              & \underline{8.8}   & 47.5  & 60.5\\
    \midrule
    + Stat. CEI         & 10.0                 & \underline{47.4}                 & 57.8 \\
    + Dyn. CEI          & \textbf{8.6}      & \textbf{46.6}     & 60.4 \\

    \bottomrule
  \end{tabular}
  \vspace{0pt}
  \label{tab:amber_results}
\end{table}

\paragraph{MMHal-Bench}
MMHal-Bench~\cite{sun_aligning_2023} targets hallucination in multimodal queries through 96 adversarially designed image-question pairs spanning 8 categories (e.g., object attributes, counting, spatial relations, adversarial objects) and 12 COCO meta-categories. Questions elicit detailed responses, evaluated via GPT-4 (augmented with image annotations and human-generated answers for text-only API compatibility), which rates for hallucinations with 94\% human agreement. GPT-4 scores the responses according to hallucination-level and informativeness. This judge-based protocol enables nuanced assessment across diverse error types, at the cost of an additional model in the loop.

As shown in \Cref{tab:mmhalbench_wide}, across the three LVLMs evaluated on MMHal-Bench, the proposed dynamic CEI consistently achieves the lowest hallucination rates, outperforming all baselines. This reduction in hallucinations is accompanied by competitive informativeness scores, yielding the highest overall performance in terms of balanced hallucination mitigation and response quality. Notably, static CEI ranks as the second-best method overall, demonstrating robust gains over prior techniques while highlighting the incremental benefits of dynamic adaptation in context embedding injection.

\begin{table}[t]\small
\centering
\caption{Performance comparison on MMHal-Bench (GPT4-evaluated) across different LVLMs. Scores (Sc~$\uparrow$) and Hallucination Rates (HR~$\downarrow$) are reported for InstructBLIP, LLaVA-1.5, and LLaVA-NeXT.}
\vspace{4pt}
\setlength{\tabcolsep}{4pt}
\begin{tabular}{l|cc|cc|cc}
\toprule
 & \multicolumn{2}{c|}{InstructBLIP} 
 & \multicolumn{2}{c|}{LLaVA-1.5} 
 & \multicolumn{2}{c}{LLaVA-NeXT} \\
Method & Sc$\uparrow$ & HR$\downarrow$ 
                & Sc$\uparrow$ & HR$\downarrow$
                & Sc$\uparrow$ & HR$\downarrow$ \\
\midrule

base        & 1.84 & 0.64   & 1.59 & 0.72   & 3.08 & 0.47 \\

+ OPERA              & 2.10 & \underline{0.58} 
                   & 2.41 & \underline{0.57} 
                   & --   & -- \\

+ VCD                & 1.75 & 0.64 
                   & 1.96 & 0.64 
                   & 2.82 & 0.57 \\

+ M3ID               & 1.70 & 0.65 
                   & 2.14 & 0.61 
                   & 2.83 & 0.57 \\

+ AvisC              & 2.03 & 0.59 
                   & 2.19 & 0.59 
                   & \underline{3.07} & \underline{0.48} \\

+ CAAC               & \textbf{2.25} & 0.64 
                   & 1.67 & 0.64 
                   & 2.92 & 0.53 \\

\midrule

+ Stat. CEI          & 2.16 & \textbf{0.57} 
                   & \textbf{2.46} & \textbf{0.56} 
                   & \textbf{3.10} & 0.49 \\

+ Dyn. CEI           & \underline{2.21} & \textbf{0.57} 
                   & \underline{2.42} & \underline{0.57} 
                   & \underline{3.07} & \textbf{0.47} \\

\bottomrule
\end{tabular}
\vspace{-10pt}
\label{tab:mmhalbench_wide}
\end{table}

\subsection{Hyperparameter Study}


We analyze the influence of the scheduler and the two key parameters, $\alpha$ and $\beta$, on hallucination reduction.

\begin{figure}[t]
  \centering
  \begin{subfigure}[b]{0.18\textwidth}
    \centering
    \includegraphics[width=\linewidth]{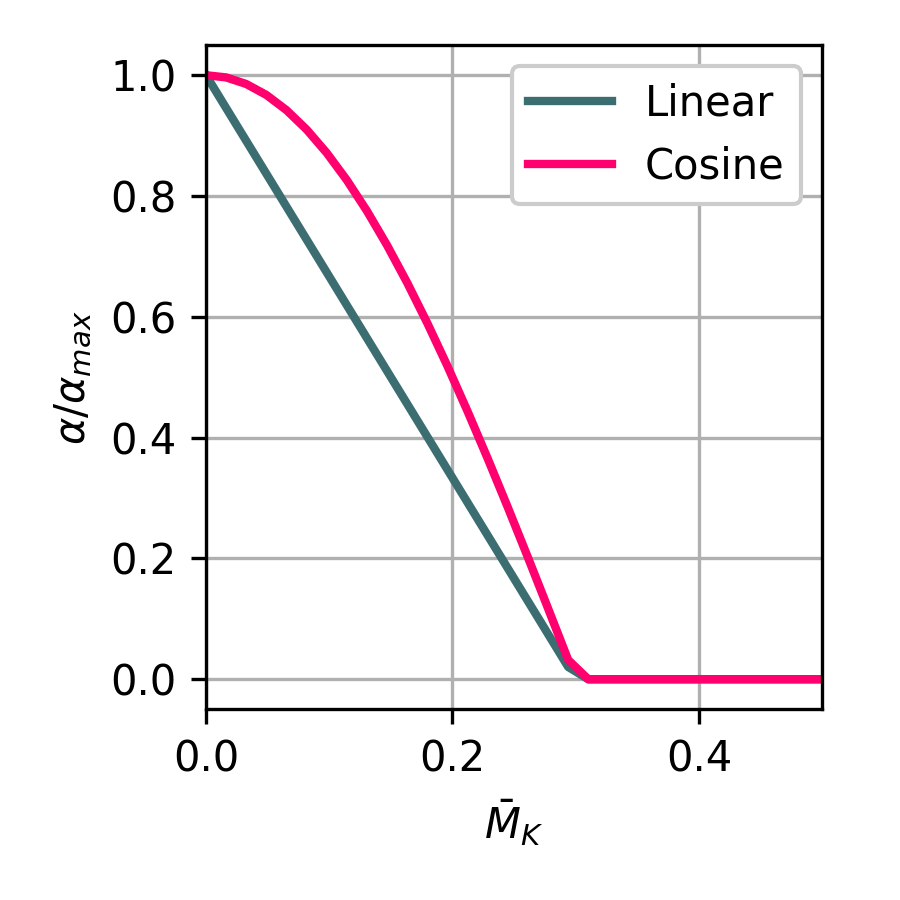}
    \caption{}
    \label{fig:scheduler}
  \end{subfigure}\hfill%
  \begin{subfigure}[b]{0.3\textwidth}
    \centering
    \includegraphics[width=\linewidth]{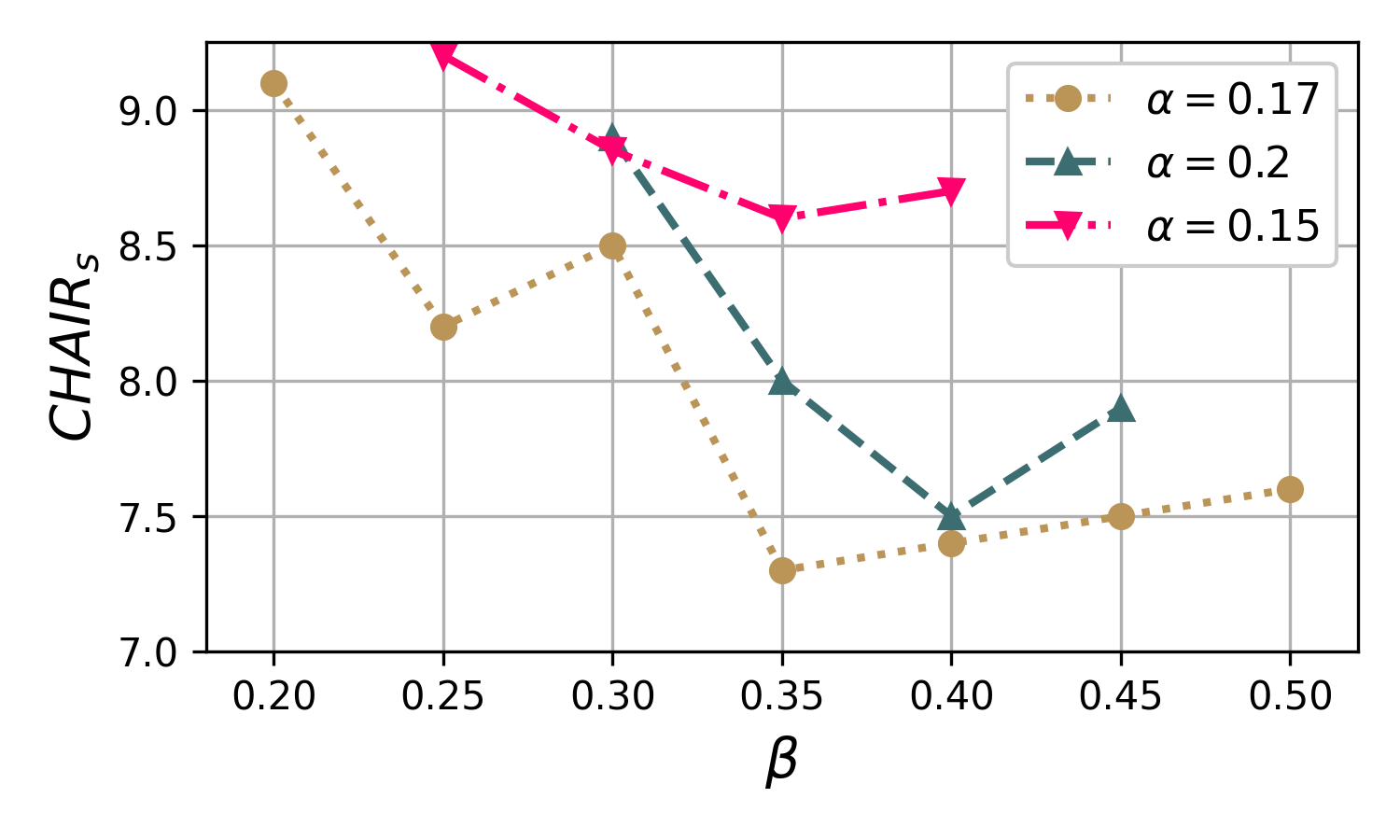}
    \caption{}
    \label{fig:hyperparam_llava_next}
  \end{subfigure}\hfill%
  \caption{(a) comparison of linear and cosine scheduler. (b) Impact of $\alpha$ and $\beta$ on CHAIR$_s$ for LLaVA-NeXT.
  }
  \vspace{-10pt}
  \label{fig:hyperparam}
\end{figure}

\Cref{fig:scheduler} compares our half-cosine mapping against a linear alternative. The half-cosine scheduler decays more gradually for low $\bar{M}_K$ values, thereby sustaining stronger corrections where commitment is weakest. In contrast, the linear scheduler declines too rapidly, attenuating interventions for low-confidence tokens and yielding elevated hallucination rates: for InstructBLIP, it produces 4.6\% higher CHAIR$_s$ and 1.7\% higher CHAIR$_i$ scores relative to half-cosine, with similar disparities observed across other LVLMs.

For cosine scheduler $\alpha$ controls the maximum injection strength. Small values under-correct, while large values produce over-correction that suppresses truthful tokens. As shown in \Cref{fig:hyperparam_llava_next}, $\alpha=0.17$ yeilds lower hallucination rates for LLaVA-NeXT compared with  $\alpha=0.15$ and  $\alpha=0.20$.

Additionally, we observe that $\beta$ determines the mean top-$K$ mass at which injection vanishes.
Too small a $\beta$ keeps injection active even during confident steps (leading to over-correction), whereas too large a value disables correction prematurely.  
For LLaVA-NeXT, $\beta \approx 0.35$--$0.40$ offers the optimal hallucination reduction (\Cref{fig:hyperparam_llava_next}).

\begin{figure}[t]
    \centering
    \includegraphics[width=\linewidth]{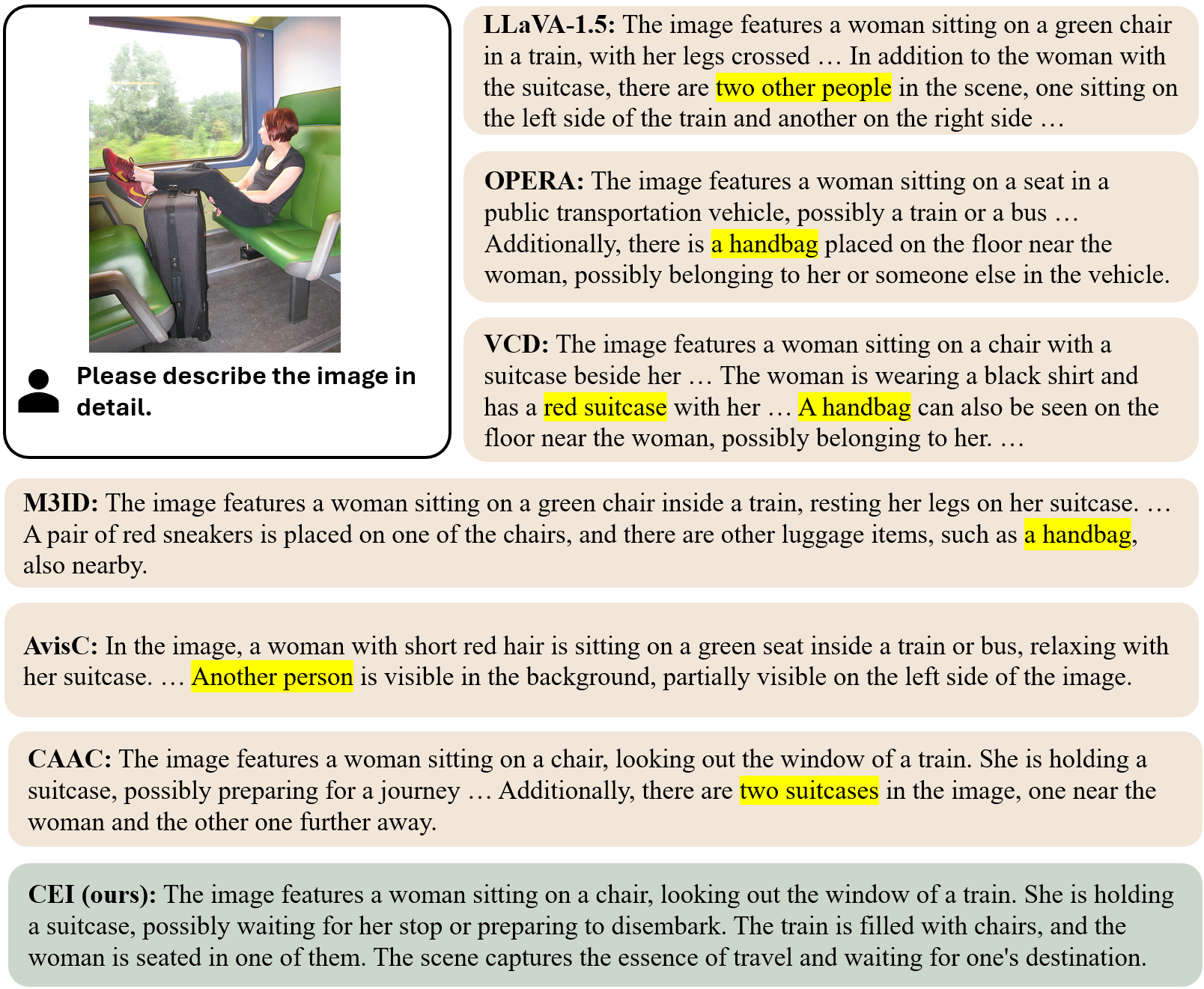}
    \caption{\textbf{Image captioning comparison.} Comparing generated captions via the baseline methods and our CEI on a sample image.}
    \label{fig:qulitative_4}
    \vspace{-10pt}
\end{figure}

\begin{figure}[t]
    \centering
    \includegraphics[width=\linewidth]{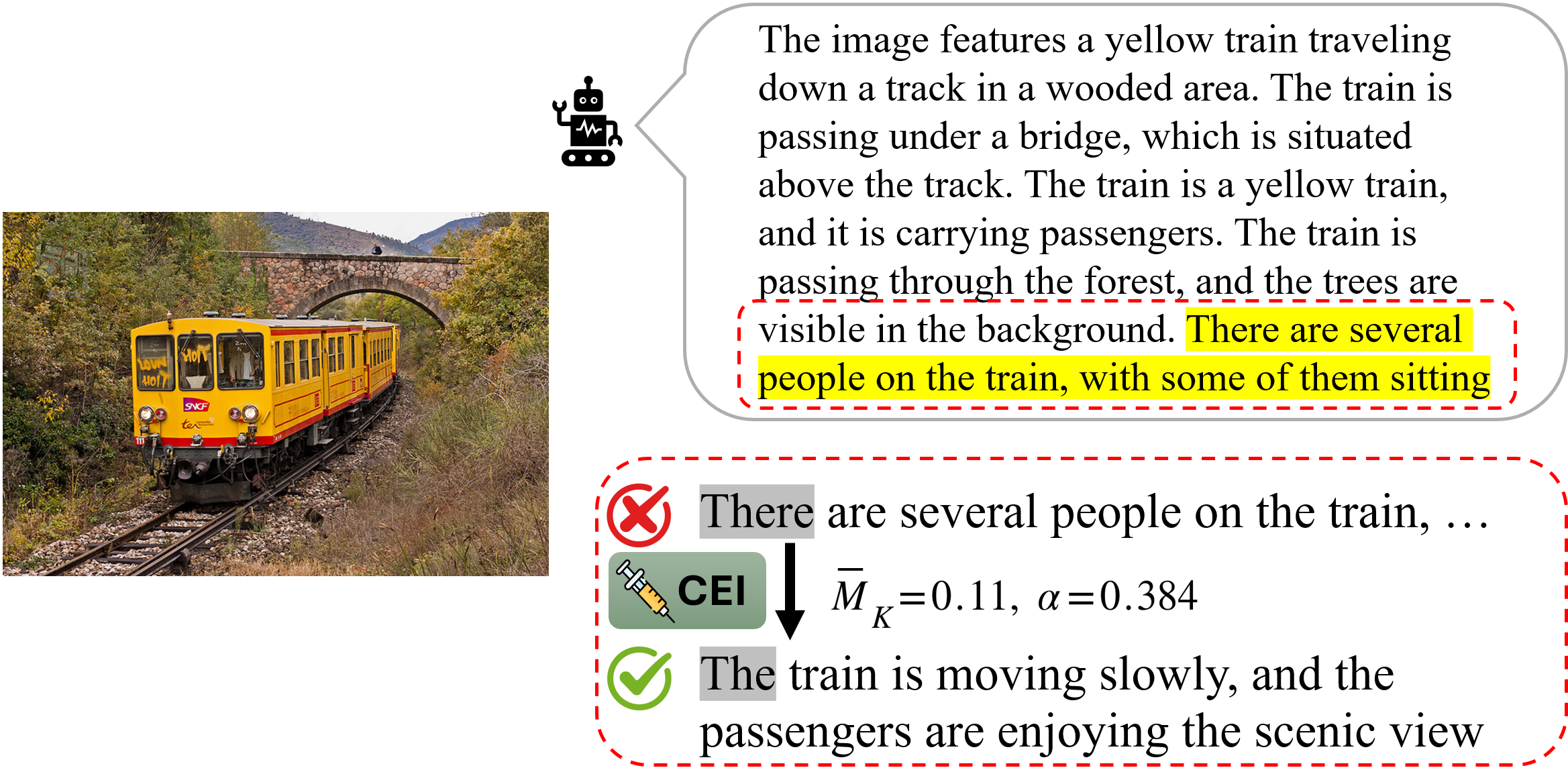}
    \caption{\textbf{Branched decoding analysis.} We generate the caption with CEI until CEI changes the top token. Then, we branch into a short greedy decoding (no CEI) to visualize how the caption would have unfolded without intervention. As shown, CEI redirects the hallucinatory continuation to a grounded one.}
    \label{fig:qualitative_type1_2}
    \vspace{-10pt}
\end{figure}

\subsection{Qualitative Evaluation}

To complement the quantitative results, we conduct a qualitative evaluation of our method by comparing its generated captions against those from baseline models on representative samples from the AMBER benchmark.  
\Cref{fig:qulitative_4} illustrates one such example: while baselines induce more hallucinations (e.g., handbag, red suitcase, two suitcases), our approach suppresses these erroneous concepts and remains faithful grounding to the image.

To better understand how CEI mitigates hallucinations, we conduct an analysis based on branched decoding. During generation under CEI, whenever the intervention alters the model's top-token selection, we temporarily branch into a short greedy decoding (without CEI) to observe how the model would have continued if CEI did not intervene. This allows us to visualize the downstream impact of a single token swap, which is often not immediately obvious at the next step alone. A key observation is that \emph{CEI most often intervenes at the onset of prospective hallucinatory phrases, preemptively redirecting the generation toward grounded continuations}. For instance, as depicted in \Cref{fig:qualitative_type1_2}, the low average top-$K$ probability mass ($\bar{M}_K = 0.11$) associated with the token ``There'' triggers a strong injection weight ($\alpha = 0.38$), swapping it for ``The'' and thereby sidestepping the ensuing hallucinatory elaboration.

Find more qualitative examples in \cref{app:qualitative_evaluation}.


\section{Conclusion}
This work bridges analysis and mitigation of hallucination in large vision–language models. Our layer-wise study revealed that truthful tokens commit earlier across decoder layers than hallucinatory ones, highlighting \emph{commitment depth} as an interpretable signal of visual faithfulness. Leveraging this finding, we proposed \emph{Context Embedding Injection (CEI)}, a lightweight, training-free intervention that continually aligns decoding with an image-conditioned context derived from the first generation step. The dynamic CEI variant further adjusts guidance strength online using the mean Top-K mass, improving robustness for long-form outputs. Across CHAIR, AMBER, and MMHal-Bench benchmarks, CEI achieves consistent reductions in hallucination while maintaining comparable coverage, demonstrating its effectiveness and generality. Future work may extend this framework to multi-image or video grounding, explore integration with contrastive decoding, and investigate commitment-aware steering for other multimodal reasoning tasks.

\section*{Limitations}
While our study demonstrates the effectiveness of CEI for reducing hallucinations in LVLMs, several limitations remain. We discuss them here to clarify the scope of our claims and to encourage future work.

\paragraph{Computational Overhead.}
Dynamic CEI introduces additional inference cost by requiring a probe forward pass at each decoding step. This results in two forward passes per token, which may limit its applicability in latency-sensitive settings. However, this overhead is comparable to that of many existing hallucination-mitigation techniques—such as contrastive decoding methods (e.g., VCD, M3ID, AvisC)—which also require duplicate forward passes. Moreover, our static variant of CEI incurs only a single additional forward pass per instance while achieving substantial improvements and outperforming most baselines across benchmarks. We primarily view dynamic CEI as a diagnostic tool for studying adaptive interventions, with runtime optimizations left for future work.

\paragraph{Evaluation Scope and Generalization.}
Our experiments focus on three widely used, COCO-style hallucination benchmarks, leaving open questions about how CEI generalizes to more specialized domains (e.g., medical imaging, autonomous driving). Although we do not evaluate on domain-specific benchmarks, MMHal-Bench includes adversarial visual reasoning examples that extend beyond simple captioning and require more complex grounding behavior. On this benchmark, CEI consistently outperforms existing mitigation methods, suggesting that the underlying mechanism is not narrowly tailored to captioning alone. Nonetheless, evaluating CEI in domain-specific settings with distinct visual distributions remains an important direction for future work.

\paragraph{Dependence on Token-Level Hallucination Labels for Analysis.}
Our preliminary insights and mechanistic analyses rely on externally provided truthful and hallucinatory token labels from annotated datasets. This dependency may limit the range of settings in which similar analyses can be performed. Importantly, these annotations are used only for analysis and do not play any role during inference or in the CEI mechanism itself; CEI remains fully unsupervised and model-agnostic at inference time. The use of annotated benchmarks is standard practice in hallucination research, particularly for probing internal model dynamics, but extending these analyses to unannotated or partially annotated settings is a promising avenue.

\paragraph{White-Box Access Requirement.}
CEI operates by modifying hidden states and requires access to internal representations and the unembedding matrix. This limits direct applicability to closed-source LVLMs accessible only through black-box APIs. However, this requirement is shared by nearly all mechanistic or decoding-time hallucination-mitigation techniques—including attention modification methods and contrastive decoding approaches. Our focus in this work is to advance understanding and control of open LVLMs, which provide the transparency necessary for reproducibility and for deeper investigation of hallucination phenomena.

\section*{Ethics Statement}
We use publicly available benchmarks and follow their licenses. Our mitigation aims to reduce factual errors; however, faithful yet harmful content remains a societal risk and should be filtered by downstream safety layers.

\section*{Potential Risks}
While our goal is to reduce hallucinations, the method does not guarantee correctness and could lead to over-trust if used in high-stakes settings. Moreover, improvements in coherence and perceived grounding may be misused to produce more convincing misleading content. We therefore position this work as a research contribution and recommend pairing it with application-specific safeguards (e.g., human oversight and domain validation) in downstream deployments.

\section*{AI Assistance Disclosure}
We used several AI language models to help edit and paraphrase text for clarity and grammar and as programming assistance. All technical content, experiments, and claims were verified by the authors.

\bibliography{references}

\appendix
\section{Additional Experimental Details}
\label{app:exp-details}

\paragraph{Models.}
We evaluate CEI on three widely used 7B-parameter LVLMs: InstructBLIP~\citep{dai_instructblip_2023}, LLaVA-1.5~\citep{liu2023visual}, and LLaVA-NeXT~\citep{liu2024llavanext}. All models are loaded via \texttt{transformers}~4.47 in 16-bit floating-point precision with no additional finetuning. CEI is applied in a model-agnostic manner by modifying hidden states at a chosen decoder layer and passing them through the original output head. For each LVLM, the injection layer and CEI hyperparameters are selected via a small grid search on held-out images.

\paragraph{Benchmarks and Metrics.}

\textbf{CHAIR.}
Following \citet{rohrbach2019object}, we use the COCO-based CHAIR setup to evaluate object hallucinations in image captioning. Ground-truth object annotations are derived from COCO labels, and generated captions are mapped to object names via lemmatization and synonym lists. We report the standard instance-level
\[
\mathrm{CHAIR}_i = \frac{\lvert \mathcal{O}_\text{hall} \rvert}{\lvert \mathcal{O}_\text{all} \rvert}
\]
where $\mathcal{O}_\text{hall}$ and $\mathcal{O}_\text{all}$ are hallucinated and all mentioned objects, and sentence-level
\[
\mathrm{CHAIR}_s = \frac{\lvert \mathcal{S}_\text{hall} \rvert}{\lvert \mathcal{S}_\text{all} \rvert}
\]
where $\mathcal{S}_\text{hall}$ is the set of captions with at least one hallucination. Lower values indicate fewer hallucinations.

\textbf{AMBER.}
AMBER~\citep{amber} is an LLM-free hallucination benchmark with dense human annotations covering object existence, hallucination targets, and salient objects. In the generative setting, models are asked to describe the image; noun phrases are extracted and matched to annotated objects and hallucination targets using the AMBER ontology. We report three metrics:
(i) CHAIR, defined as the average fraction of hallucinated object mentions per response;
(ii) Hal, the proportion of responses with any hallucination;
and (iii) Cover, which measures response informativeness as the fraction of annotated objects correctly mentioned in each caption, averaged over all examples. Thus, lower CHAIR/Hal and higher Cover indicate better grounding.

\textbf{MMHal-Bench.}
MMHal-Bench~\citep{sun_aligning_2023} consists of adversarial visual reasoning questions designed to elicit hallucinations under challenging prompts. Responses are scored on a discrete 0–5 scale by a factually augmented evaluation pipeline. We follow the official protocol and report
\[
\text{Score} = \frac{1}{N} \sum_{n=1}^{N} s_n
\]
and hallucination rate
\[
\text{HalRate} = \frac{1}{N} \sum_{n=1}^{N} \mathbb{I}[s_n < 3],
\]
where $s_n$ is the score for question $n$ and $N$ is the total number of questions. Higher Score and lower HalRate correspond to better performance. For all benchmarks, we allow up to 512 new tokens to avoid truncating long generations, as hallucinations frequently occur in later decoding steps.

\paragraph{Baseline Methods.}
We compare CEI with five training-free hallucination mitigation methods:

\begin{itemize}
    \item \textbf{OPERA}~\citep{huang_opera_nodate} modifies beam search by penalizing over-confident beams and reallocating probability mass based on a retrospection signal. This encourages generations that remain consistent with visual evidence rather than language priors. We use the official implementation with the authors’ recommended hyperparameters for each LVLM.
    
    \item \textbf{VCD}~\citep{vcd} (Visual Contrastive Decoding) constructs contrastive candidate paths by perturbing the visual input and comparing logits between original and perturbed runs. Tokens whose probability is not robust to visual changes are down-weighted during decoding, reducing visually unsupported continuations. We adopt the configuration provided in the AvisC repository.
    
    \item \textbf{AvisC}~\citep{avisc} extends contrastive decoding by introducing adaptive penalties that focus on visually grounded paths. It dynamically adjusts the contrastive strength during decoding to discourage hallucinations while preserving fluency. We use the authors’ recommended hyperparameters for each model.
    
    \item \textbf{M3ID}~\citep{m3id} uses multi-modal contrastive signals to identify and suppress tokens that are weakly supported by visual features. By contrasting predictions under different visual conditions, it targets hallucinations that arise from over-reliance on language priors. We follow the default settings in the official codebase.
    
    \item \textbf{CAAC}~\citep{fazli_mitigating_2025} (Confidence-Aware Attention Calibration) calibrates self-attention maps to re-balance attention to image tokens based on token-level confidence. It aims to counteract the skew where the decoder ignores visual features in favor of language statistics.
\end{itemize}

All baselines are evaluated using their official implementations, and we retain the decoding parameters (e.g., temperature, top-$p$) recommended in the original papers to ensure fair comparison.

\paragraph{CEI Hyperparameters and Grid Search.}
For dynamic CEI, we tune three hyperparameters per LVLM: the maximum injection weight $\alpha_{\max}$, the cutoff parameter $\beta$ in the mean top-$K$ mass scheduler, and the injection layer. We perform a coarse grid search over a small set of layers and a few candidate values of $\alpha_{\max}$ and $\beta$ spanning low, medium, and high intervention regimes. The final settings are:
\begin{itemize}
    \item \textbf{InstructBLIP}: $\alpha_{\max} = 0.40$, $\beta = 0.70$, injection layer $= 10$.
    \item \textbf{LLaVA-1.5}: $\alpha_{\max} = 0.25$, $\beta = 0.55$, injection layer $= 10$.
    \item \textbf{LLaVA-NeXT}: $\alpha_{\max} = 0.17$, $\beta = 0.35$, injection layer $= 10$.
\end{itemize}
Static CEI uses the same injection layer but replaces the dynamic schedule with a fixed injection weight derived from the average effective $\alpha$ in the dynamic setting.

\paragraph{Hardware and Runtime.}
All experiments are run on a single server with 4$\times$NVIDIA H100 40\,GB GPUs and 512\,GB system RAM. We parallelize evaluation across images and models. A full AMBER run (512-token limit) takes roughly 12 hours for InstructBLIP and 10 hours for LLaVA-1.5, while CHAIR and MMHal-Bench evaluations are somewhat faster due to smaller dataset sizes. Runtime differences are mainly driven by model architecture and average caption length.

\paragraph{Reproducibility.}
We fix random seeds, use consistent preprocessing and prompt templates across models, and share decoding parameters across baselines and CEI variants. We plan to release code, evaluation scripts, and configuration files (including all CEI hyperparameters and grid-search ranges) to facilitate reproducibility.

\section{Context Embedding as a Grounding Signal}
\label{app:context-embedding-analysis}

To substantiate the hypothesis that the hidden state of the final prompt token---hereafter the \emph{context embedding}---serves as a semantic grounding signal, we conduct an intrinsic analysis measuring its alignment with truthful and hallucinatory tokens. This embedding corresponds to the final-layer hidden state of the last input position (image~+~query) in the decoder, which is directly unembedded to produce the logits of the first generated token. The experiment aims to determine whether this representation is inherently more aligned with truthful words that describe the image faithfully, thereby justifying its use as a grounding signal in our Context Embedding Injection (CEI) framework.

\begin{figure}[t]
  \centering
  \begin{subfigure}[b]{0.8\linewidth}
    \centering
    \includegraphics[width=\linewidth]{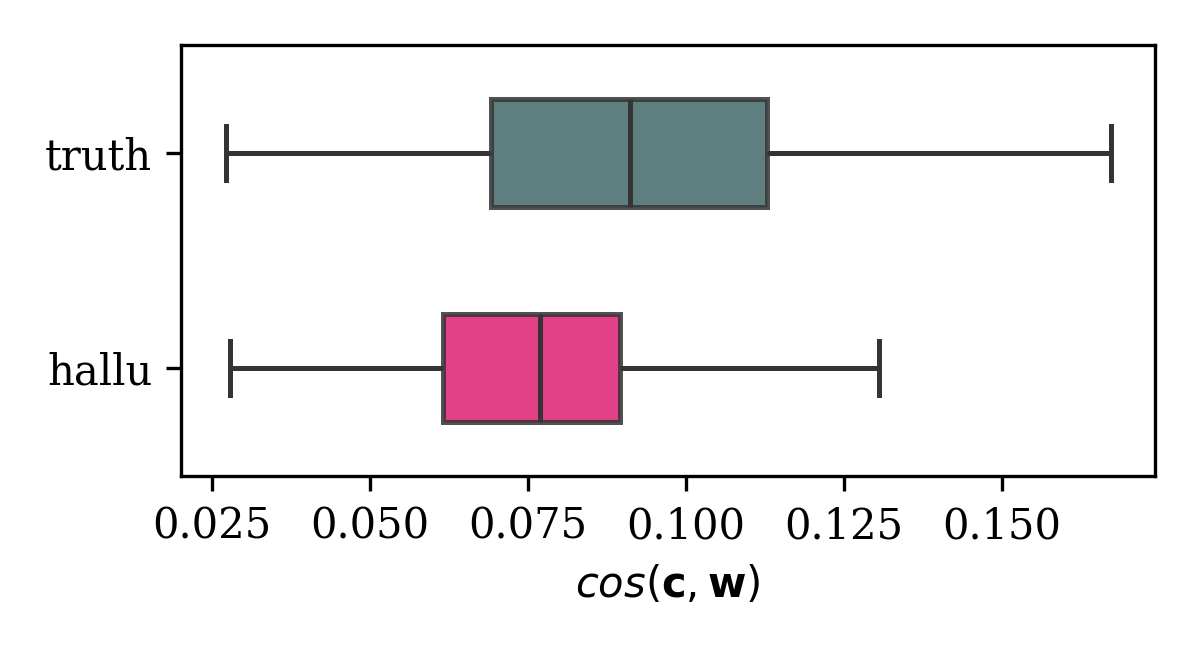}
    \caption{}
    \label{}
  \end{subfigure}\hfill%
  \begin{subfigure}[b]{0.8\linewidth}
    \centering
    \includegraphics[width=\linewidth]{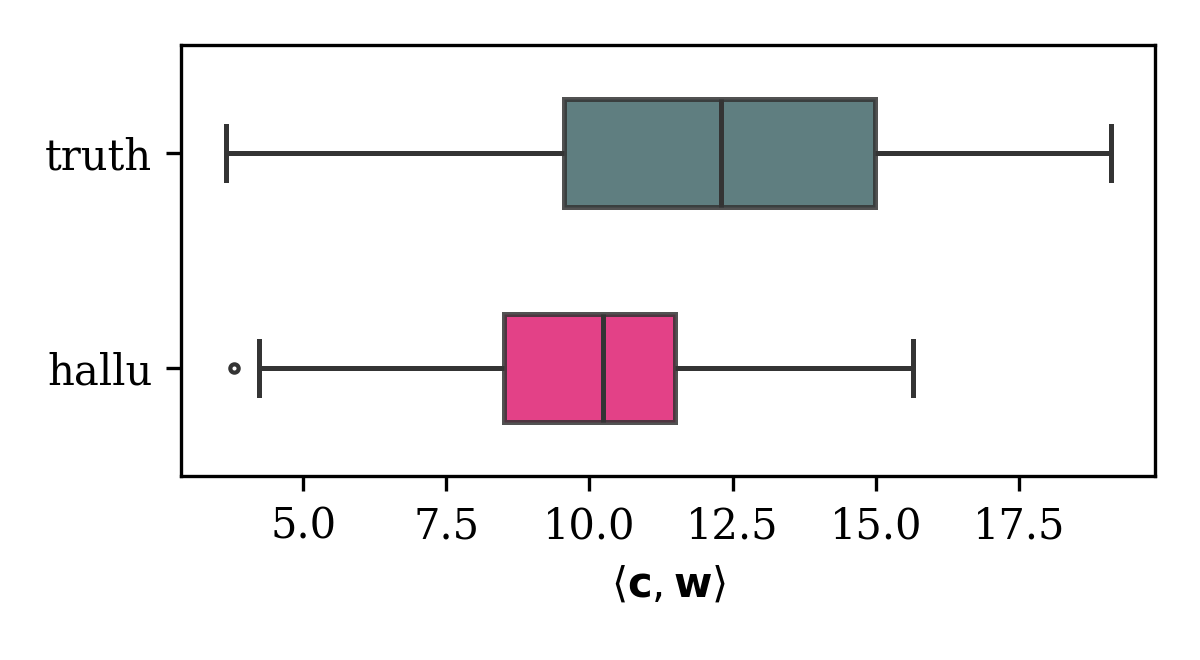}
    \caption{}
    \label{}
  \end{subfigure}\hfill%
  \caption{Box plot of (a) cosine similarities $\cos(\mathbf{c}, \mathbf{w})$, and (b) dot product $\langle \mathbf{c}, \mathbf{w} \rangle$ between the context embedding $\mathbf{c}$ and target token embedding $\mathbf{w}$. Blue and orange boxes denote truthful and hallucinatory tokens respectively.
  }
  \label{fig:cosine-and-dot-similarities}
\end{figure}

\paragraph{Experimental setup.}
We perform the analysis using captions generated by InstructBLIP on the AMBER benchmark~\citep{amber}, which provides human-annotated word-level labels distinguishing truthful and hallucinatory objects. Each generated caption is tokenized, and the annotations are aligned to the corresponding subword tokens. For multi-token words, we average the representations across their constituent sub-tokens. The representation of a token \(\mathbf{w}\) is taken from the unembedding matrix of the language head (\(W_U\)), i.e., the column vector used to map hidden states to logits for token \(y\). This choice ensures that all similarity measures are computed in the same space the model uses for generation, where the context embedding \(\mathbf{c}\) naturally resides.

\paragraph{Metrics.}
We report three complementary measures of alignment: (1) the raw dot product \(\langle \mathbf{c}, \mathbf{w} \rangle\), which reflects the model’s true logit-space affinity for each token; (2) the raw cosine similarity \(\cos(\mathbf{c}, \mathbf{w}) = \frac{\mathbf{c}^\top \mathbf{w}}{\|\mathbf{c}\|\,\|\mathbf{w}\|}\), capturing purely directional alignment; and (3) the centered cosine similarity \(\cos(\mathbf{c}-\mu, \mathbf{w}-\mu)\), where \(\mu\) is the mean token embedding vector across the vocabulary. The latter mitigates anisotropy in transformer embedding spaces, following the post-processing approach of \citet{muennighoff_mteb_2023}. These complementary metrics allow us to disentangle whether truthful tokens align with \(\mathbf{c}\) due to directional consistency, overall magnitude, or both.

\paragraph{Results.}
Across all metrics, truthful tokens exhibit stronger alignment with the context embedding than hallucinatory tokens. As shown in Figure~\ref{fig:centered-cosine}, the centered cosine similarities of truthful tokens are substantially higher, and the mean difference between the two classes is significantly positive (95\% CI excluding zero). Similar trends hold for the raw cosine and dot-product measures (\Cref{fig:cosine-and-dot-similarities}), demonstrating that the context embedding both points toward and activates truthful token directions in the model’s output space. These findings provide direct empirical evidence that the context embedding is semantically grounded and not an arbitrary choice: it naturally encodes visually faithful information, aligning closely with truthful tokens. This supports its role as a grounding signal for CEI and aligns with prior observations that early decoding representations encode truthfulness cues~\citep{zhao2024first}.

\begin{figure}[t]
  \centering
  \begin{subfigure}[b]{0.9\linewidth}
    \centering
    \includegraphics[width=\linewidth]{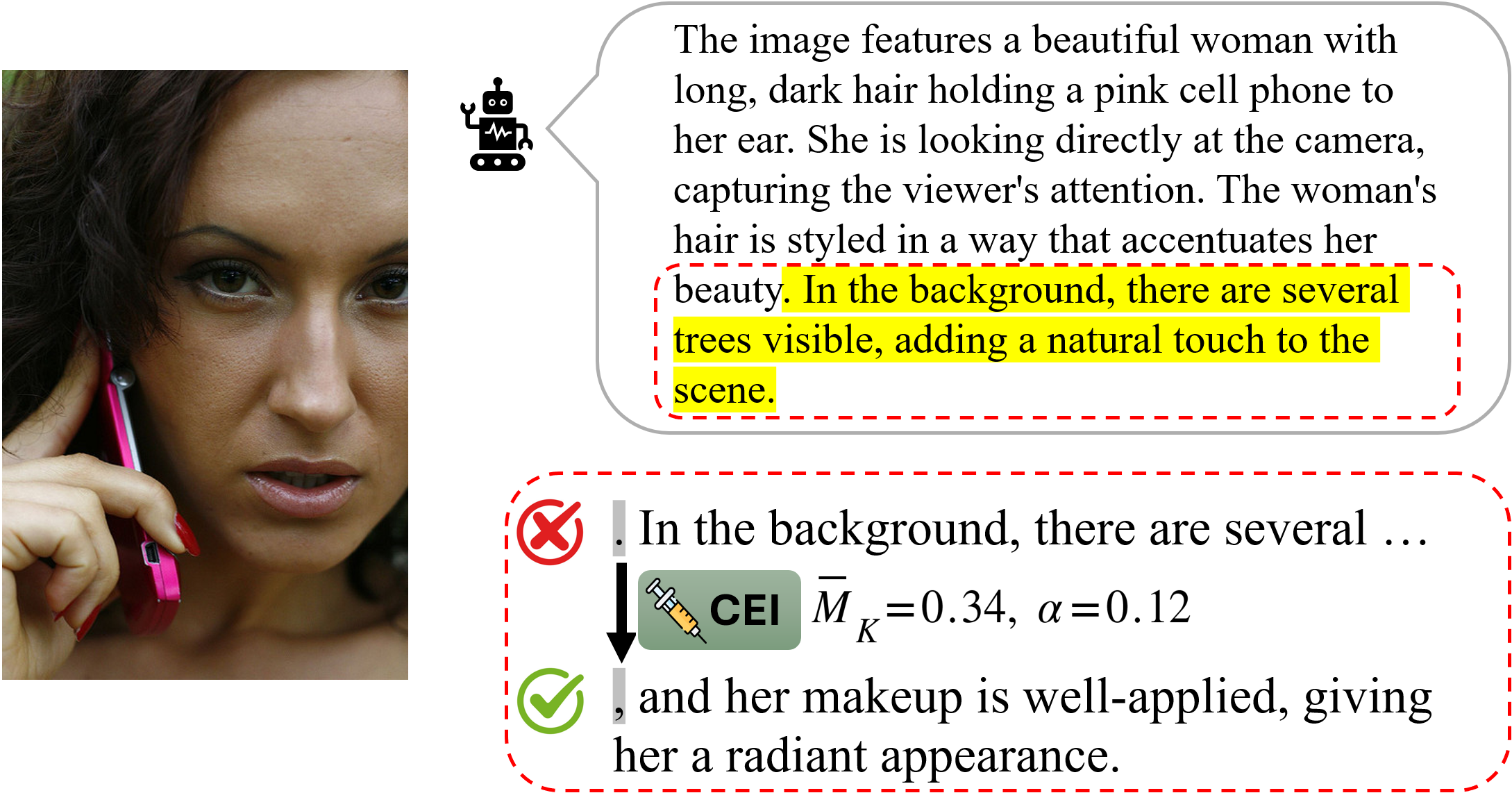}
    \caption{}
  \end{subfigure}\hfill%
  \begin{subfigure}[b]{0.9\linewidth}
    \centering
    \includegraphics[width=\linewidth]{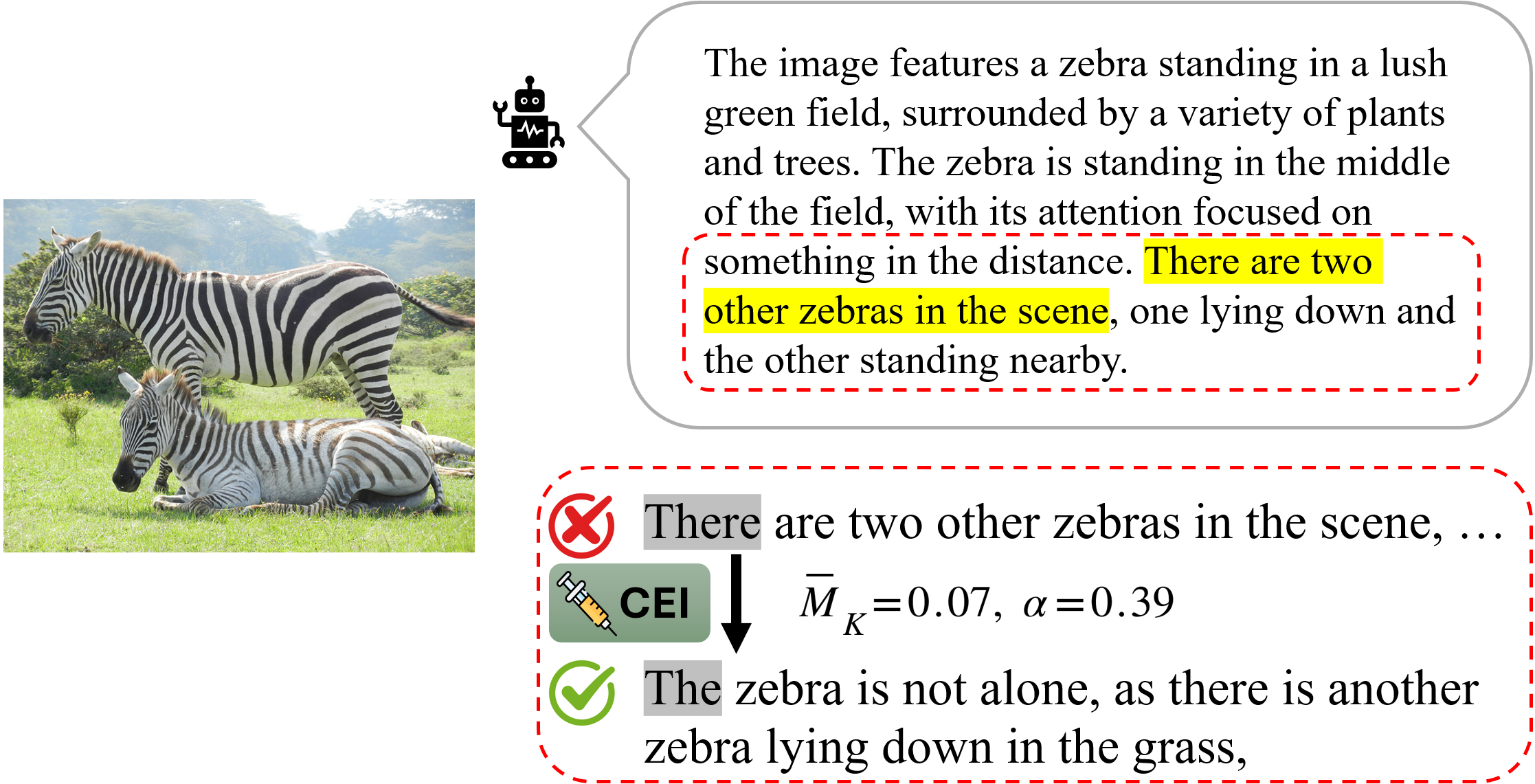}
    \caption{}
  \end{subfigure}\hfill%
  \caption{Examples showing how injection eliminates a hallucinatory statement and realigns the generated sequence with the input image. The captions are generated with InstructBLIP and once a token swap occurs we branch into a greedy decoding for 20 tokens to visualize how the sequence would have unfolded hadn't CEI intervened.
  }
  \label{fig:qualitative-type1}
\end{figure}

\begin{figure*}[t]
  \centering
  \begin{minipage}{0.49\textwidth}
    \centering
    \begin{subfigure}{\linewidth}
      \centering
      \includegraphics[width=\linewidth]{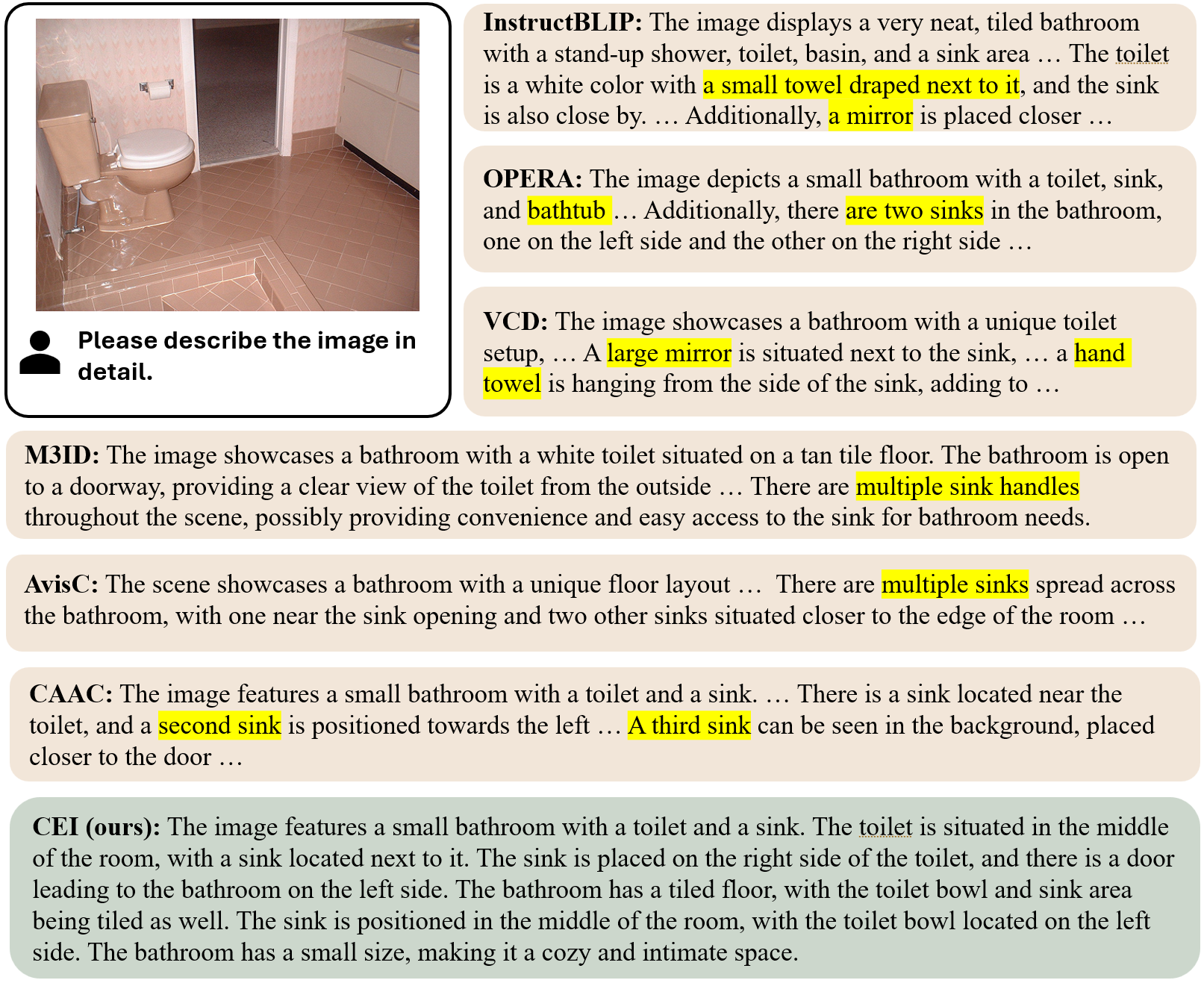}
      \caption{}
    \end{subfigure}
    \\
    \begin{subfigure}{\linewidth}
      \centering
      \includegraphics[width=\linewidth]{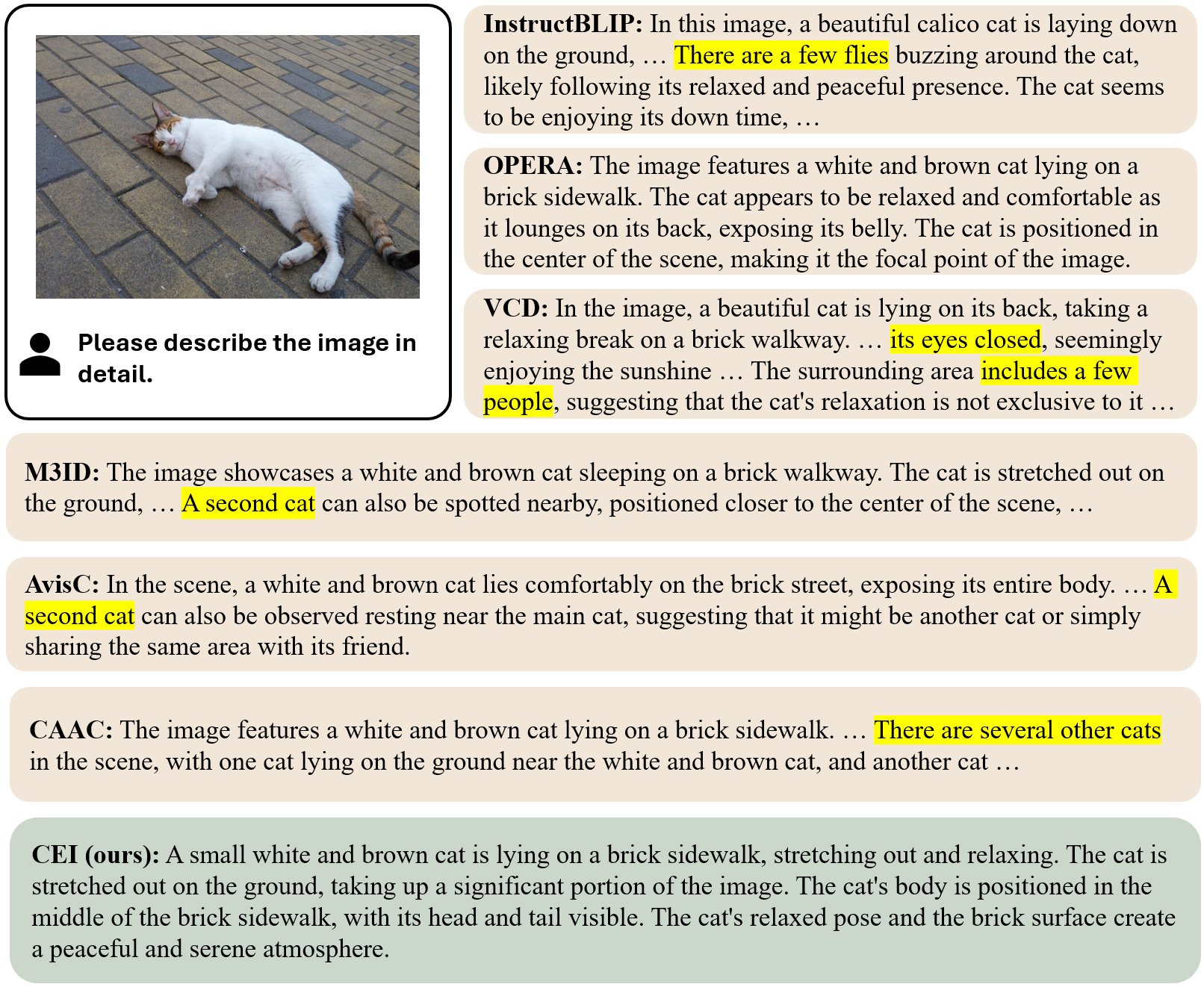}
      \caption{}
    \end{subfigure}
    \\
    \begin{subfigure}{\linewidth}
      \centering
      \includegraphics[width=\linewidth]{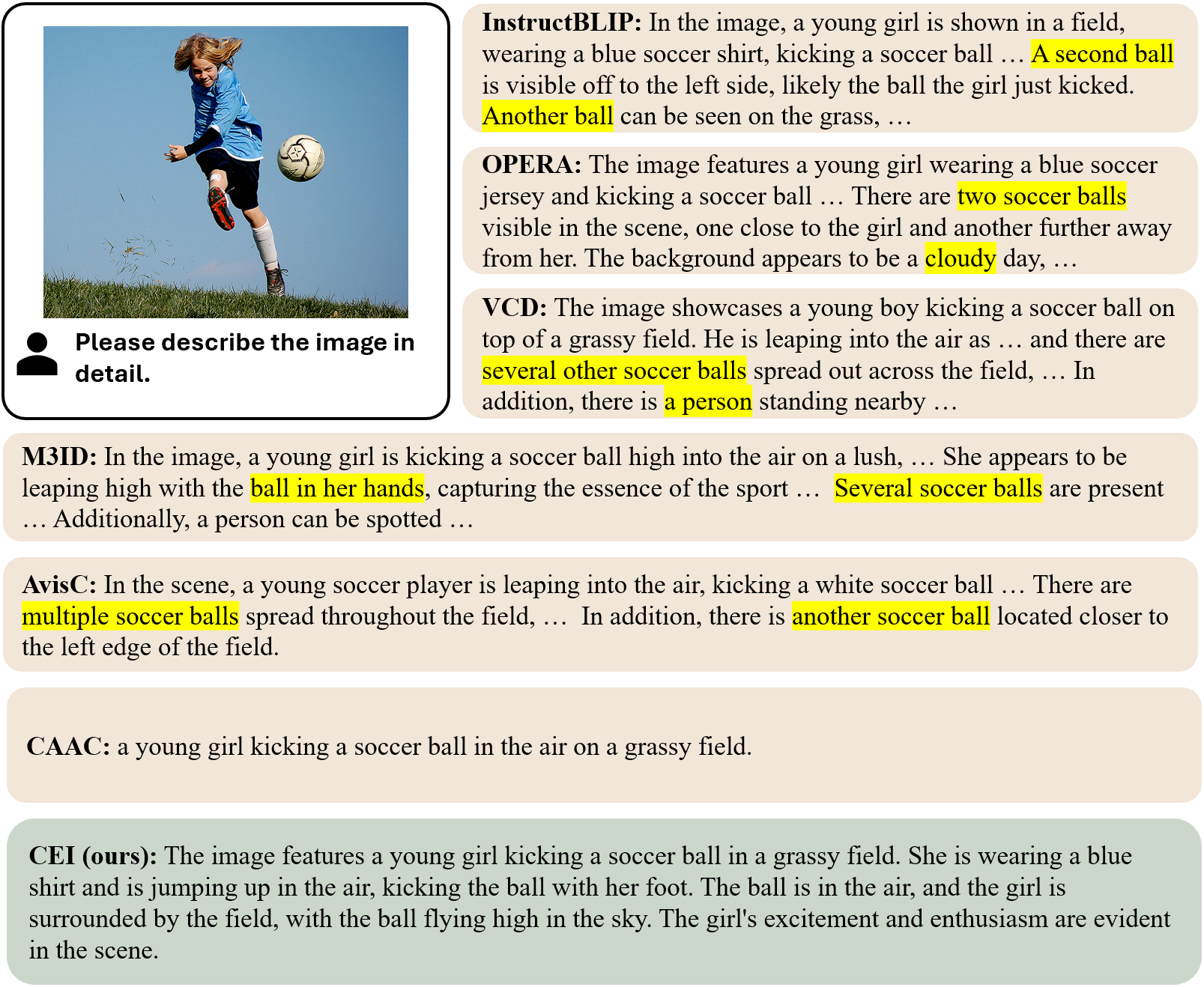}
      \caption{}
    \end{subfigure}
    \caption{Comparison of CEI outputs with baseline methods for the InstructBLIP model. Hallucinations are highlighted in yellow for easy comparison. MaxTokens is set to 512 for all models.}
    \label{fig:InstructBLIP-qualitative_type_2}
  \end{minipage}
  \hfill
  \begin{minipage}{0.49\textwidth}
    \centering
    \begin{subfigure}{\linewidth}
      \centering
      \includegraphics[width=\linewidth]{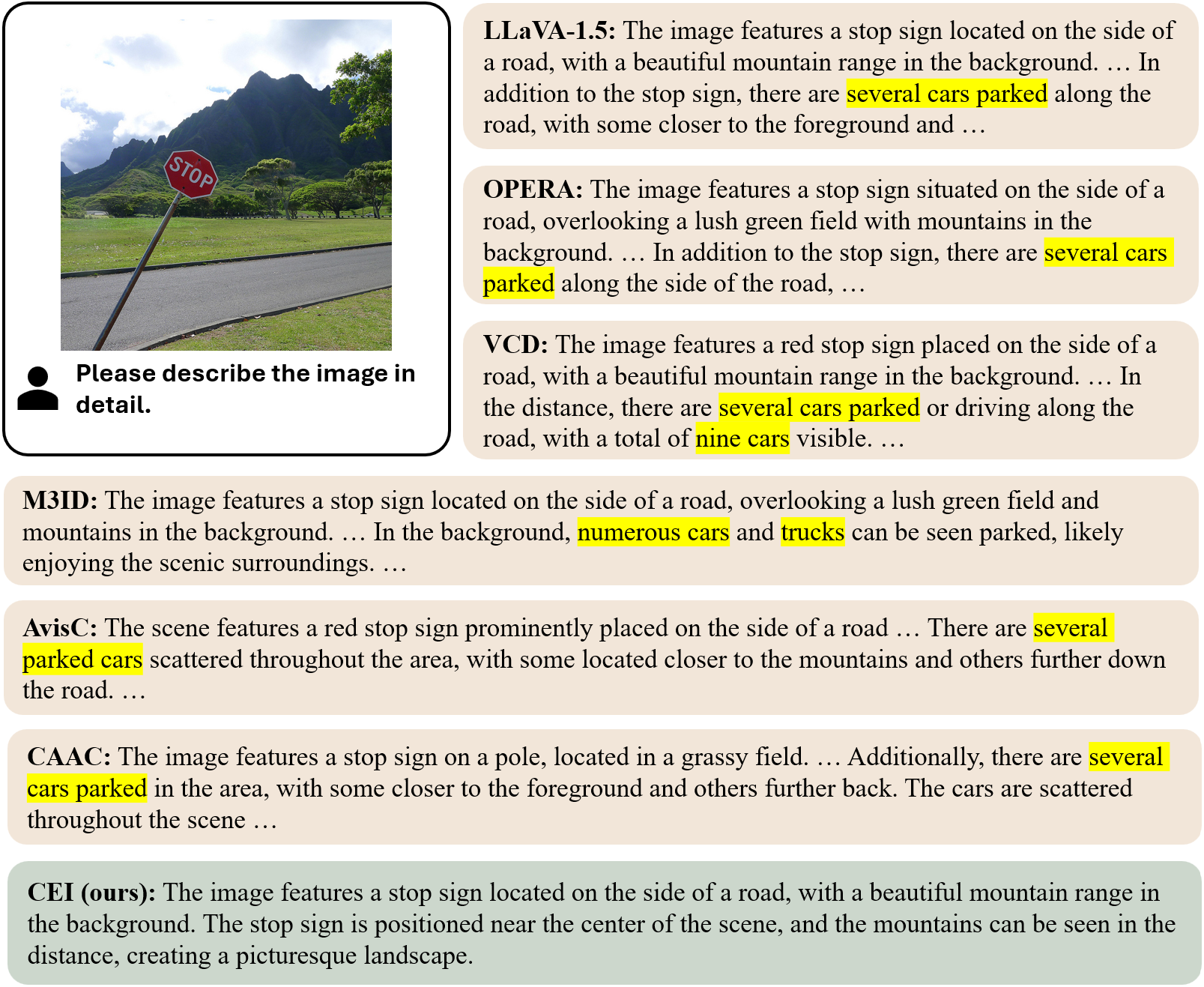}
      \caption{}
    \end{subfigure}
    \\
    \begin{subfigure}{\linewidth}
      \centering
      \includegraphics[width=\linewidth]{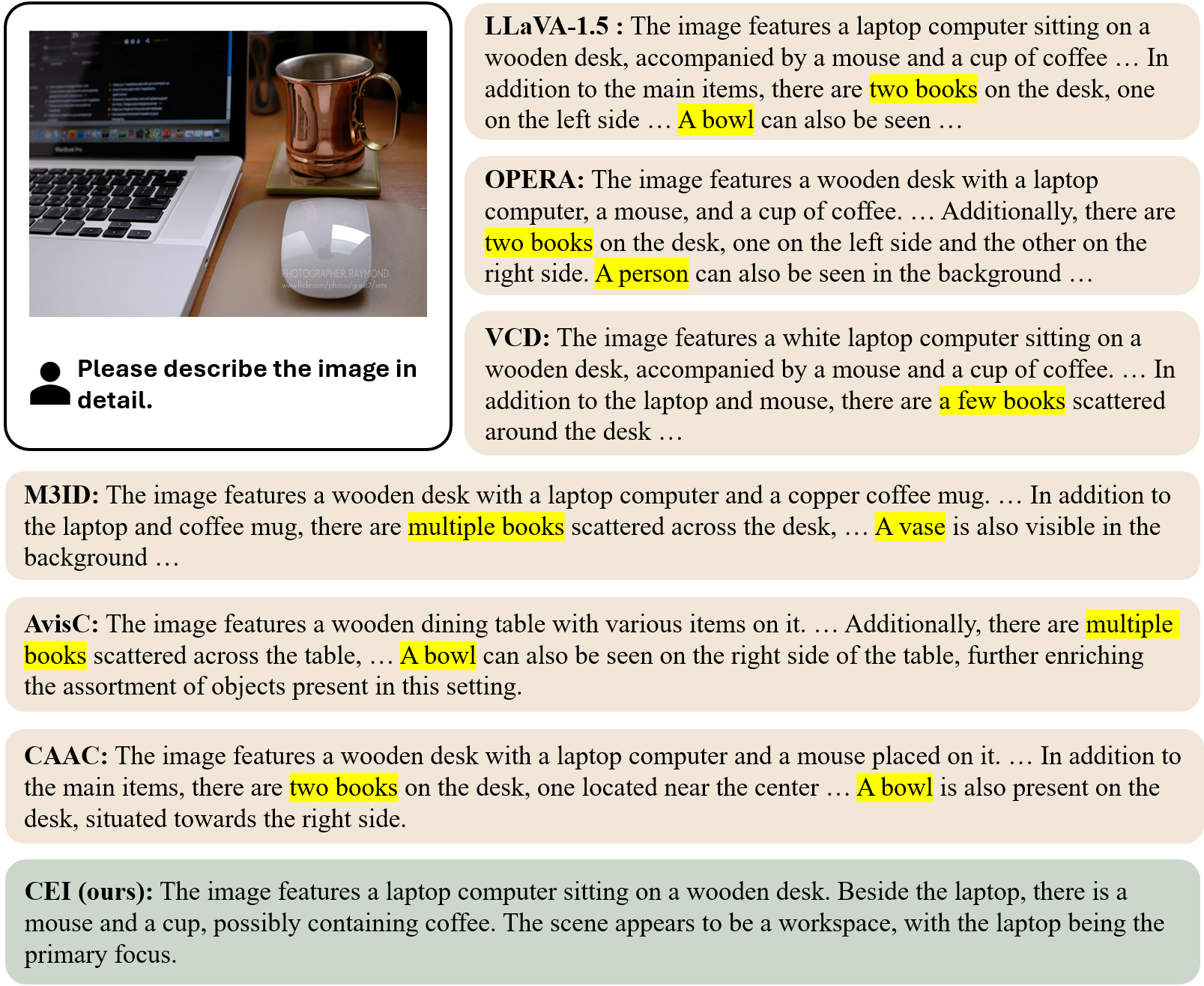}
      \caption{}
    \end{subfigure}
    \\
    \begin{subfigure}{\linewidth}
      \centering
      \includegraphics[width=\linewidth]{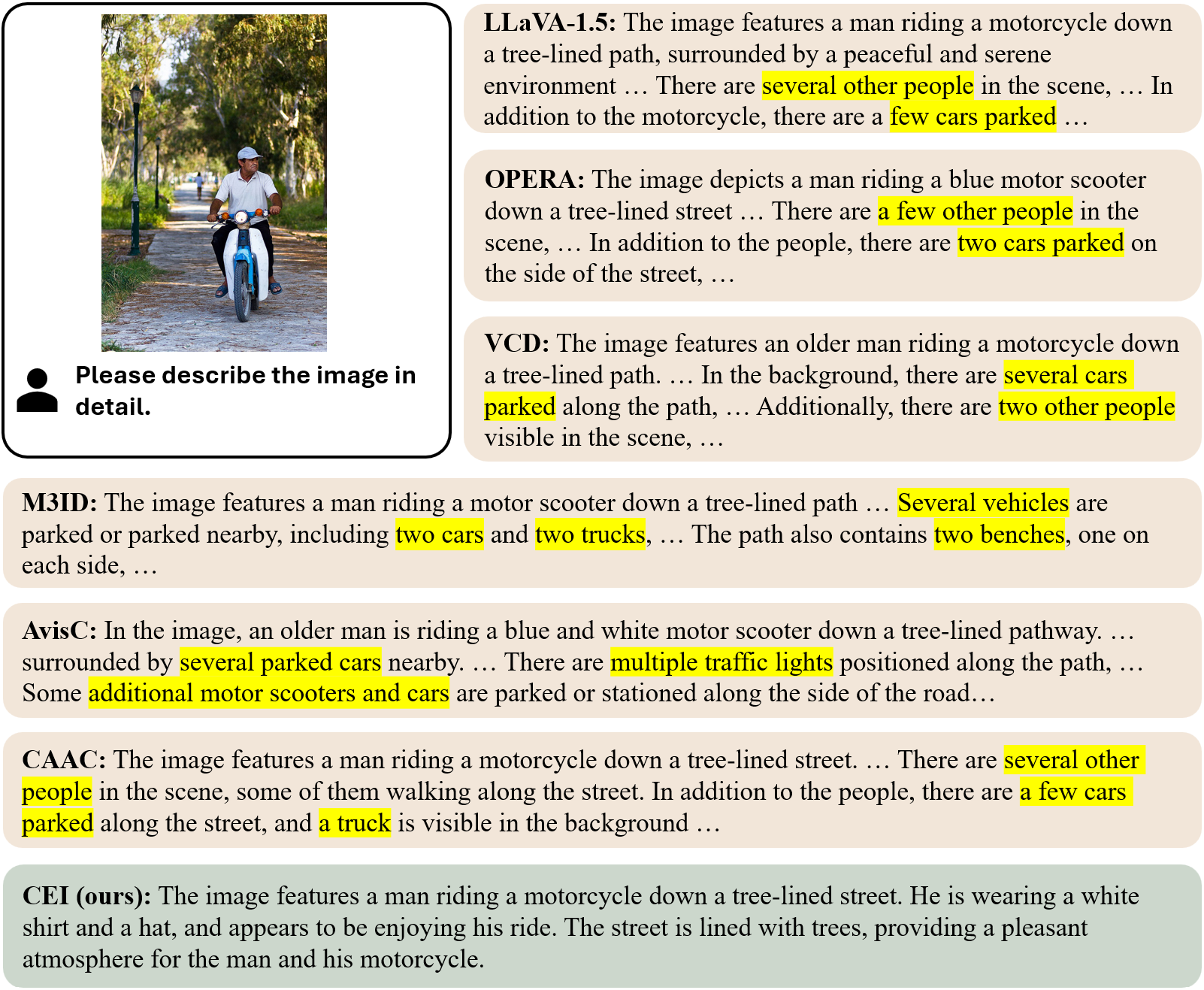}
      \caption{}
    \end{subfigure}
    \caption{Comparison of CEI outputs with baseline methods for the LLaVA-1.5 model. Hallucinations are highlighted in yellow for easy comparison. MaxTokens is set to 512 for all models.}
    \label{fig:LLaVA-qualitative_type_2}
  \end{minipage}
\end{figure*}

\section{Additional Qualitative Examples}
\label{app:qualitative_evaluation}

In this section, we provide an extended set of qualitative examples to further illustrate the behavior of our method across a diverse range of images from the AMBER benchmark. For each example, we display the input image alongside captions generated by the baseline models and by our approach. Hallucinatory tokens are highlighted for ease of comparison.

Across examples, we consistently observe the same pattern noted in the main text: baseline models often fail to mitigate hallucinations in long generated sequences and sometimes even introduce objects, attributes, or relations that are not visually grounded. In contrast, our method reliably suppresses these hallucinations while preserving the correct visual details.

We include a variety of challenging scenarios---cluttered scenes, small objects, atypical viewpoints, and visually ambiguous contexts (\Cref{fig:InstructBLIP-qualitative_type_2}, \Cref{fig:LLaVA-qualitative_type_2}).

To further elucidate the intervention mechanism of CEI, we also include examples of branched decoding analyses here (\Cref{fig:qualitative-type1}). As described in the main text, these analyses involve generating captions under active CEI influence until a top-token swap occurs (identified by contrasting the first forward pass with no injection and second forward pass), at which point we temporarily suspend the intervention to pursue a short ``greedy'' decoding path (without CEI) for 5--10 subsequent tokens. This branching reveals counterfactual trajectories, highlighting how CEI redirects generation away from hallucinatory continuations toward faithful descriptions. A recurring observation is that interventions typically occur at the onset of prospective hallucinatory phrases, preemptively averting downstream errors.

\section{Related Work}
\label{app:related_work}
\subsection{Large Vision-Language Models}
Large vision-language models (LVLMs) represent a significant advancement in multimodal AI, extending the capabilities of large language models (LLMs) by integrating visual processing. Typically, LVLMs consist of a vision encoder (e.g., CLIP~\citep{radford_learning_2021}, ViT~\citep{dosovitskiy_image_2021}) to extract image features, an alignment module such as a linear projection~\cite{liu2023visual, liu2024llavanext} or Q-former~\citep{dai_instructblip_2023, minigpt4} to map these features into the language model's embedding space, and an LLM backbone (e.g., LLaMA~\citep{touvron_llama_2023}, Vicuna~\cite{zheng_judging_2023}) for autoregressive generation conditioned on both visual and textual inputs. This architecture enables LVLMs to handle diverse tasks, including image captioning, visual question answering (VQA), object detection, and multi-modal reasoning~\citep{chen_vlp_2023, wu_multimodal_2023}, by concatenating visual tokens with text embeddings for unified processing. Recent families~\citep{liu2024llavanext, wang_qwen2-vl_2024, ye_mplug-owl_2024} scale model and data while improving visual tokenization (dynamic resolution, multi-scale inputs~\citep{liu2024llavanext, wang_qwen2-vl_2024}) and positional fusion (e.g., M-RoPE~\citep{wang_qwen2-vl_2024}), enabling broad capability gains across captioning, VQA, and free-form assistance. However, LVLMs are prone to hallucination which undermines their reliability in safety critical applications~\citep{bai_hallucination_2025}.

\subsection{Hallucination Mitigation in LVLMs}
Hallucinations in large vision-language models (LVLMs) refer to the generation of content that deviates from the visual input, such as fabricating non-existent objects, attributes, or relations. They are often attributed to over-reliance on linguistic priors, training data biases, or modality misalignment~\citep{bai_hallucination_2025, pope, liu_survey_2024}. Mitigation approaches mainly fall into three categories: fine-tuning~\citep{gunjal_detecting_2024, jiang_hallucination_2024, kim_exposing_2023}, post-hoc correction~\citep{yin_woodpecker_2023, zhou_analyzing_2024}, and decoding-time techniques~\citep{vcd, opera, suo_octopus_2025, fazli_mitigating_2025, an_mitigating_2025, yang_nullu_2025}.

Decoding-time methods stand out for their efficiency and applicability without retraining. Contrastive decoding (CD)~\citep{li_contrastive_2023, chuang_dola_2024} is prominent, contrasting original and perturbed inputs to favor visual fidelity. Examples include Visual Contrastive Decoding (VCD)~\citep{vcd} using image noise, Multi-Modal Mutual Information Decoding (M3ID)~\citep{m3id} via query masking, and Image-Biased Decoding (IBD)~\citep{zhu_ibd_2024} adjusting probabilities from image-biased models. Attention-centric approaches calibrate cross-modal interactions—e.g., Assembly of Global and Local Attention (AGLA)~\citep{an_mitigating_2025} combines attention maps to reduce biases, and Confidence-Aware Attention Calibration (CAAC)~\citep{fazli_mitigating_2025} adaptively amplifies attention to image tokens. In contrast, our method operates directly in the embedding space by continually aligning the token representations with a visual grounding signal.

\end{document}